\documentclass[sigconf, 10pt]{acmart}

\AtBeginDocument{%
  }

\setcopyright{acmcopyright}
\copyrightyear{2018}
\acmYear{2018}
\acmDOI{XXXXXXX.XXXXXXX}

\acmConference[Conference acronym 'XX]{Make sure to enter the correct
  conference title from your rights confirmation emai}{June 03--05,
  2018}{Woodstock, NY}
\acmPrice{15.00}
\acmISBN{978-1-4503-XXXX-X/18/06}

\usepackage[normalem]{ulem}
\usepackage{pifont}
\usepackage{multirow}
\usepackage{xspace} 
\usepackage{mathrsfs}
\usepackage{amsfonts}
\usepackage{amsmath}
\usepackage{enumitem}
\usepackage{graphicx}
\usepackage{tabularray}
\usepackage{subfig}
\usepackage{listings}

\newcommand{\sysname}{LUT-NN\xspace}

\newcommand{\etal}{\hbox{\emph{et al.}}\xspace}

\newcommand{\add}[1]{{{{#1}}}}

\begin{document}

\title{
\sysname: Empower Efficient Neural Network Inference with Centroid Learning and Table Lookup}

\author{Xiaohu Tang} %
\authornote{Contribution during internship at Microsoft Research}
\affiliation{
  \institution{State Key Lab of CAD\&CG, Zhejiang University\\
  Microsoft Research}
  \country{}
}
\email{tigertang@zju.edu.cn}

\author{Yang Wang} 
\affiliation{
  \institution{Microsoft Research}
  \country{}
}
\email{Yang.Wang92@microsoft.com}

\author{Ting Cao} %
\authornote{Corresponding author}
\affiliation{
  \institution{Microsoft Research}
  \country{}
}
\email{ting.cao@microsoft.com}

\author{Li Lyna Zhang}
\affiliation{
  \institution{Microsoft Research}
  \country{}
}
\email{lzhani@microsoft.com}

\author{Qi Chen}
\affiliation{
  \institution{Microsoft Research}
  \country{}
}
\email{cheqi@microsoft.com}

\author{Deng Cai}
\affiliation{
  \institution{State Key Lab of CAD\&CG, Zhejiang University}
  \country{}
}
\email{dengcai@cad.zju.edu.cn}

\author{Yunxin Liu}
\affiliation{
  \institution{Institute for AI Industry Research (AIR), Tsinghua University}
  \country{}
}
\email{liuyunxin@air.tsinghua.edu.cn}

\author{Mao Yang}
\affiliation{
  \institution{Microsoft Research}
  \country{}
}
\email{maoyang@microsoft.com}

\renewcommand{\shortauthors}{Xiaohu, Yang, Ting et al.}

\begin{abstract}
On-device Deep Neural Network (DNN) inference consumes significant computing resources and development efforts. To alleviate that, we propose \sysname, the first system to empower inference by table lookup, %
to reduce inference cost. %
\sysname learns the typical features for each operator, named centroid, and precompute the results for these centroids to save in lookup tables. During inference, the results of the closest centroids with the inputs can be read directly from the table, as the approximated outputs without computations.  

\sysname integrates two major novel techniques: (1) differentiable centroid learning through backpropagation, which adapts three levels of approximation to minimize the accuracy impact by centroids; (2) table lookup inference execution, which comprehensively considers different levels of parallelism, memory access reduction, and dedicated hardware units for optimal performance. \sysname is evaluated on multiple real tasks, covering image and speech recognition, and nature language processing. %
Compared to related work, %
\sysname improves accuracy by 66\% to 92\%, achieving similar level with the original models. \sysname reduces the cost at all dimensions, including FLOPs ($\leq16\times$), model size ($\leq7\times$), latency ($\leq 6.8\times$), memory ($\leq6.5\times$), and power ($\leq41.7\%$).    

\end{abstract}

\vspace{-1in}



\acmYear{2023}\copyrightyear{2023}
\setcopyright{acmlicensed}
\acmConference[ACM MobiCom '23]{The 29th Annual International Conference on Mobile Computing and Networking}{October 2--6, 2023}{Madrid, Spain}
\acmBooktitle{The 29th Annual International Conference on Mobile Computing and Networking (ACM MobiCom '23), October 2--6, 2023, Madrid, Spain}
\acmPrice{15.00}
\acmDOI{10.1145/3570361.3613285}
\acmISBN{978-1-4503-9990-6/23/10}

\maketitle

\vspace{-0.1in}
\section{Introduction}

DNNs are widely used in mobile applications, offering users unparalleled intelligent services. However, DNNs are computation hungry workloads, mainly composed of linear computation operators, heavily stressing the limited hardware resources on mobile devices.   

As a result, huge efforts have been taken for efficient and affordable DNN inference on mobile devices, such as model compression~\cite{blalock2020state, gou2021knowledge}, sophisticated computing operator optimization~\cite{onnx, tensorflow, MACE, wang2021asymo}, tensor compilers for operator generation~\cite{chen2018tvm,liang2022romou}, and customized DNN accelerators~\cite{JiaoHL20, tpuv4, nnpi}. These methods require redesigning or re-implementing model structures, computation operators, or accelerators repeatedly for diverse deployment scenarios.

Different from these directions, this paper explores a new possibility to replace computation operators of DNNs, for reduced inference cost, as well as the tedious efforts of operator development. To achieve this, we first rethink the essence of inference. %
Each layer of a DNN model is to output another level of features given the input features. For example, the front layers of an image recognition model output low-level features (e.g., edges and lines), and subsequent layers output high-level features (e.g., faces and objects). The fact is that the features of different images for each layer have semantic similarity. %
A vivid example is that though cats and horses are different input images for a model, their ear features result in similar output for the layer that extracts it. Same for language tasks, similar words of different sentences could have similar embeddings and output results.   

Based on this similarity, the question is that whether the typical features %
can be learned for each computation operator of DNNs, so that the output of these typical features can represent the output of diverse features. %
If so, by precomputing and saving the output of typical features, the output of future inputs can be read directly without computation. 

\begin{figure}[!htb]
    \centering
    \includegraphics[width=0.8\columnwidth]{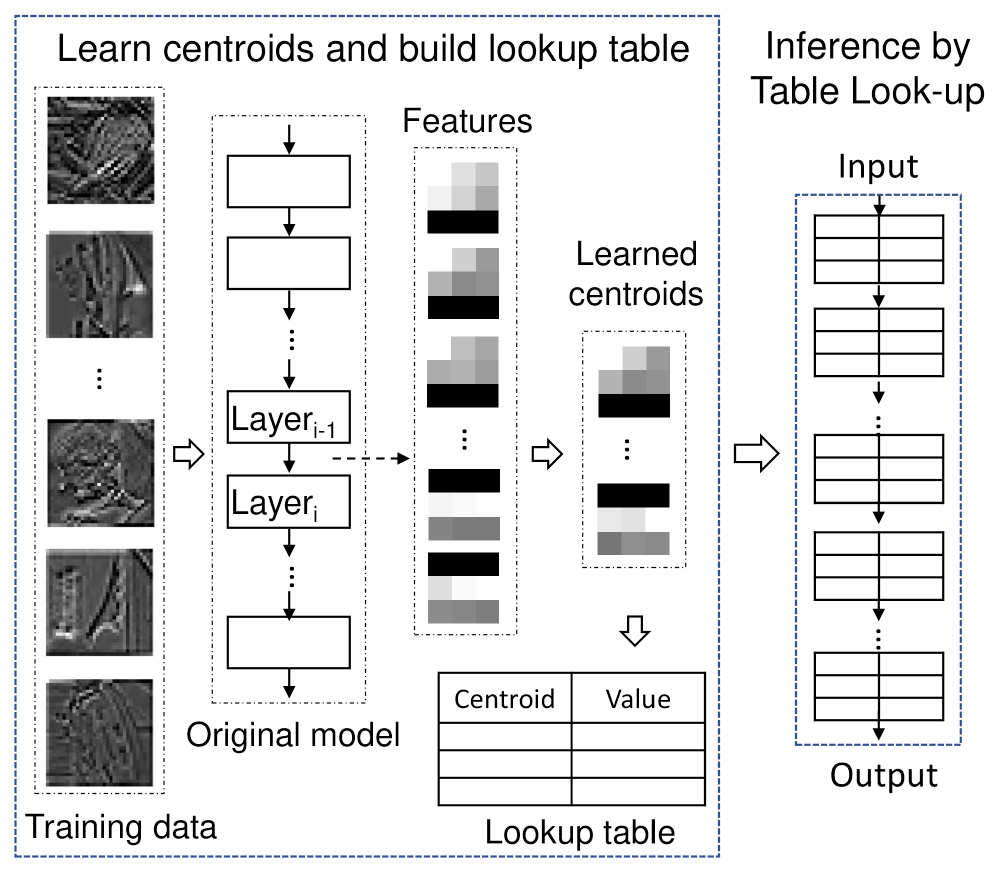}
    \caption{\sysname transforms model linear-computation layers to table lookup for inference. %
    }
    \vspace{-0.1in}
    \label{fig:highlevel}
\end{figure}

Towards answering this question, we propose \sysname, a novel system to empower DNN inference by table lookup. The system learns the typical features, named \emph{centroid}, for each linear computation operator, and precompute the results for these centroids to save in lookup tables. During inference, the results of the closest centroids with the input features can be read directly from the table as the approximated output of this operator. The required computation cost for inference is significantly reduced. Fig.~\ref{fig:highlevel} shows an overview of \sysname using real data samples of a model. \sysname integrates two major techniques: (1) differentiable centroid learning through backpropagation, and (2) table-lookup inference execution.

\textsc{Maddness}~\cite{MMWM} initiates the centroid learning for matrix multiplication (MM), leveraging the technique of Product Quantization (PQ). PQ is an effective vector quantization algorithm widely used for dataset compression~\cite{jegou2010product,GuoKCS15}. It compresses the dataset by clustering the vectors in the dataset first, and then learning the centroids to represent vectors in each cluster. By using PQ, \textsc{Maddness} learns the centroids from the training dataset for a single-operator image classifier, and precomputes the MM results for the centroids, as the approximation for future input matrices. 

However, as we will show in the paper, directly applying \textsc{Maddness} to the computation operators of a DNN model results in poor accuracy (80\% accuracy drops for CIFAR-10). A similar conclusion is also from McCarter \etal~\cite{McCarter2022}.
We first analyze the issue and expose the main reason for the poor result is \emph{the different optimization goal of centroid learning for PQ and DNNs.} The goal of PQ is to learn the centroids that can minimize the total error/distance between centroids and vectors, while the goal of DNNs is to minimize the final loss function. The two goals have no direct relationship. Without considering the loss function, errors introduced by centroids in \textsc{Maddness} are accumulated layer by layer, resulting in poor accuracy, making the resulting DNN model unreliable and undeployable in real-world applications.

Therefore, we identify the key factor for DNN centroid learning is to \emph{pass the model loss to each operator through backpropagation}, and iteratively adjust the centroids by the gradients to minimize the model loss. However, the challenge is that currently it is not possible to learn centroids through backpropogation, since encoding a vector to the closest centroid e.g., \verb|argmin| or hashing in \textsc{Maddness}, is not differentiable, and thus the gradients cannot be calculated.  %

To solve this challenge, we propose the novel \textbf{differentiable centroid learning} technique for DNN. Through backpropagation, it adapts three different levels of approximation introduced by centroids to model inference by three methods, and minimize the accuracy loss. (1) To enable gradient calculation and iteratively adjust centroids, \sysname uses \emph{soft-PQ} method. It uses the continuous approximation of \verb|argmax|, i.e., \verb|softmax|, for backward pass. The forward pass still uses \verb|argmin| for model loss calculation, since \verb|argmin| will be used for inference. Soft-PQ can adapt the approximation introduced by centroids to reduce accuracy loss. (2) Since the use of \verb|softmax| introduces approximation to \verb|argmin|, \sysname uses \emph{learned temperature} method, to learn the hyperparameter \verb|temperature| of \verb|softmax| for each operator, to tradeoff accuracy loss and learning convergence. (3) Since the lookup tables for precomputed results are the main cost for \sysname, scalar quantization is used to reduce the table size. To adapt the approximation introduced by this scalar quantization, \sysname uses \emph{quantization-aware training}, which uses quantized tables in the forward pass and real-value tables in the backward pass, to reduce accuracy loss. By the techniques above, the centroids can be learned, which empowers table lookup to replace the diverse DNN linear computations while maintain similar-level of model accuracy.   %

\sysname achieves a high FLOPs reduction (up to 16$\times$). To run \sysname on resource-limited devices, we design and implement the \textbf{table lookup inference execution} on commodity ARM and x86 CPUs. The challenge is that table lookup is a new inference paradigm, and no direct support from current systems and accelerators. Our design comprehensively considers different levels of hardware parallelism, memory access reduction, and dedicated hardware units, to fully utilize hardware resources. Specifically, we align the size of codebook and lookup table to the width of SIMD instructions for data-level parallelism, interleave the instructions of \verb|argmin| reduction for instruction-level parallelism, reside the codebook in an inner loop for cache locality, and leverage the shuffle instruction for fast table access.

We evaluate \sysname on a range of tasks, including image recognition tasks:  CIFAR~\cite{cifar}, GTSRB~\cite{GTSRB}, SVHN~\cite{SVHN}, and ImageNet~\cite{imagenet}; speech recognition task: Google Speech Command~\cite{speech_command}; nature language processing (NLP) tasks: GLUE~\cite{glue} and \add{a regression task: UTKFace~\cite{zhifei2017cvpr} age prediction}, and cover both CNN and BERT models. For all these tasks, compared to directly applying \textsc{Maddness}, \sysname can improve the accuracy by 66\% to 92\%. The accuracy difference between \sysname and the original DNN models ranges from +1.98\% (Speech Command) to -2.42\% (ImageNet).   

For inference efficiency, \sysname can outperform baselines in all dimensions. 
The selected baselines include both the state-of-the-art hand-written inference system ONNX Runtime~\cite{onnx}, and also the well-tuned TVM~\cite{chen2018tvm} generated kernels as the performance upper-bound of linear computations. The detailed results include: FLOPs reduces $5.7\times\sim16\times$; model (or disk) size reduces $3.4\times\sim7\times$; end-to-end inference latency reduces $1.3\times\sim6.8\times$; memory usage reduces $1.4\times\sim6.5\times$; and peak power reduces $15\%\sim41.7\%$. 
The real speedup is less than the FLOPs reduction due to the unfriendly support for table lookup in hardware. A customized table-lookup unit or accelerator could unleash the full potential of \sysname. 

\sysname takes the first trial towards this new table-lookup inference paradigm, which potentially much simplifies the inference system and hardware design and improves efficiency.  %
To sum up, the key contributions of this paper are as follows. \add{The code is open sourced at \url{https://github.com/lutnn}. } 
\begin{itemize}
    \item \sysname is the first to empower DNN inference by table lookup to reduce inference cost.  
    \item It can achieve similar level of model accuracy by the novel differentiable centroid learning technique for DNNs through backpropagation. 
    \item It designs the new table lookup operation to enable this new inference paradigm on commodity CPUs.
    \item We implement the learning and inference pipelines. Superior results are shown on all evaluated dimensions. %
\end{itemize}

\vspace{-0.15in}
\section{Background and Motivation}

Each data sample in the training set can be assumed as a vector. This section will first introduce the concept of vector quantization and its efficient solution, PQ, and then show the poor results of directly applying PQ to DNN inference. 

\vspace{-0.05in}
\subsection{Product Quantization: Background}
\label{sec:pq}
\vspace{-0.05in}

Quantization has been well studied in information theory~\cite{gray1998quantization}. It reduces the cardinality of a dataset by using \emph{centroids} to represent the data samples. The set of centroids is called a \emph{codebook}. For vector quantization, the dataset is composed of $D$-dimension vectors. The vector quantizer can encode each vector to a centroid in the codebook. %

\begin{figure}[!htb]
    \centering
    \vspace{-0.15in}
    \includegraphics[width=0.8\linewidth]{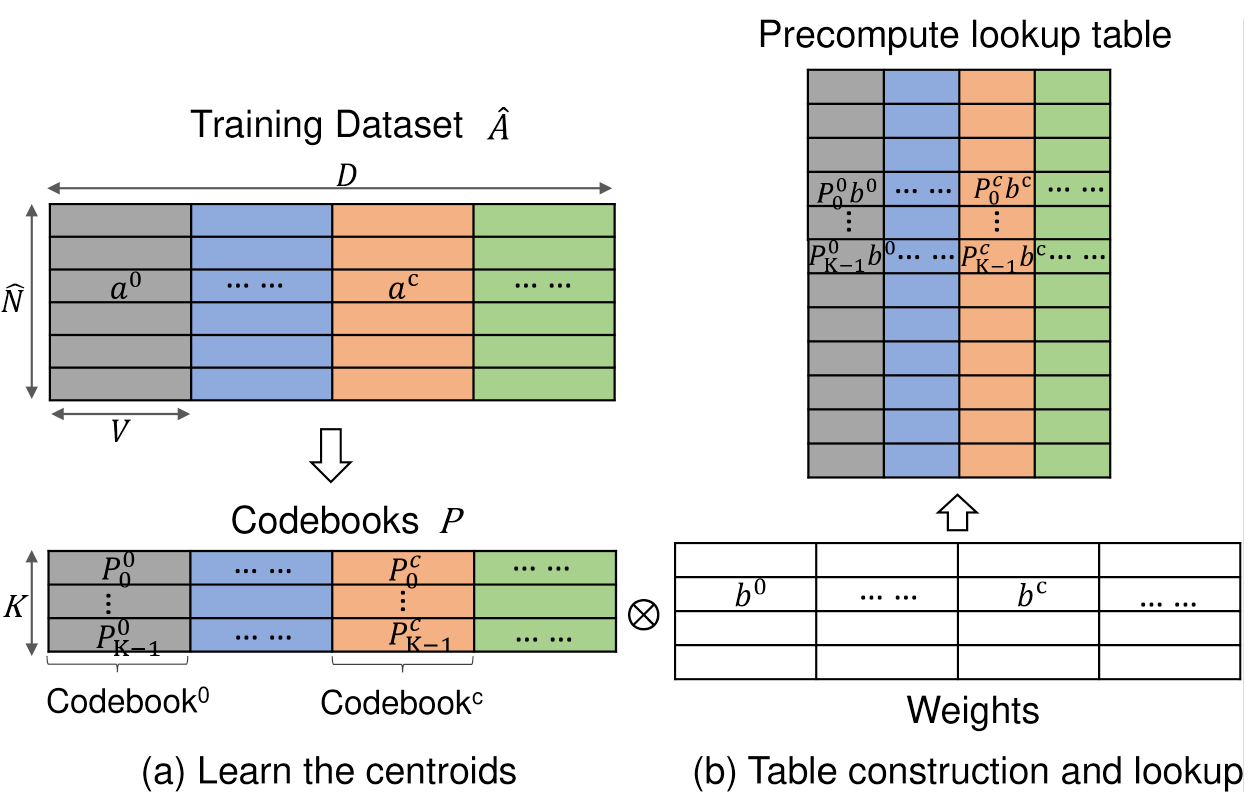}
    \vspace{-0.1in}
    \caption{Product Quantization for AMM. Each color marks a sub-vector dataset.}
    \vspace{-0.25in}
    \label{fig:product_quantization}
\end{figure}

To reduce the cost of vector quantization, PQ is proposed. The essence of it is to decompose the high-dimensional vector space into the Cartesian product of sub-vector spaces and then quantize these sub-vector spaces separately. As shown in Fig.~\ref{fig:product_quantization}(a), it splits the $D$-dimension vector into $C$ distinct  $V$-dimension sub-vectors ($D=C \cdot V$). The sub-vectors are quantized separately using $C$ codebooks. The quantization result of the vector is the concatenation of the $C$ centroids.

\label{sec:background_product_quantization}

\textbf{Centroid learning} We now formulize the PQ process. To quantize a $D$-dimension vector $a\in \mathbb{R}^{D}$, PQ needs to learn the centroids from a training dataset $\hat{A}\in \mathbb{R}^{\hat{N}\times D}$ composed of vectors with the same distribution as $a$. PQ first decomposes the vectors in the dataset into $C$ distinct $V$-dimension sub-vectors, notated as $\hat{A}^c\in \mathbb{R}^{\hat{N}\times V}$ (marked in different colors in Fig.~\ref{fig:product_quantization}). To make optimal quantization, the centroid learning process is to find the $K$ centroids $P^{c}$ (i.e., the $c^{th}$ codebook) for $\hat{A}^c$ by $k$-means~\cite{Lloyd82}, which can minimize the distance sum of each sub-vector $\hat{A}^{c}_{i}$ and its nearest centroid $P^{c}_{k}$, as shown in Eq.~\ref{eq:pqgoal}.
\vspace{-0.1in}
\begin{equation}
\underset{P}{\mathrm{arg\ min}}\sum_{c}\sum_{i}\left \| \hat{A}^{c}_{i}-P^{c}_{k} \right \|^2
\label{eq:pqgoal}
\end{equation}
\vspace{-0.1in}

\textbf{Sub-vector encoding} With the learned centroids $P$, for an input vector $a$, PQ can encode it as the concatenation of the nearest centroids for each sub-vector. The encoding function for a sub-vector is shown in Eq.~\ref{eq:encoding}.  By this vector decomposition method, the centroids can represent $K^C$ different vectors by only $K\times C$ memory cost.   
\vspace{-0.05in}
\begin{equation}
g^{c}(a^{c})=\underset{k}{\mathrm{arg\ min}}\left \| a^{c}-P^{c}_{k} \right \|^2
\label{eq:encoding}
\end{equation}
\vspace{-0.1in}

\textbf{Hashing for acceleration with bigger error } Learning centroids is a NP-hard problem. Vanilla PQ uses $k$-means to learn centroids and encode sub-vectors. $k$-means satisfies Lloyd optimal conditions~\cite{Lloyd82,jegou2010product} and can get a local optimal quantization error. However, the $k$-means encoding costs high to compute the Euclidean distance for each sub-vector with each centroid, as shown in Eq.~\ref{eq:encoding}.

To reduce the encoding cost, some works propose hashing methods to encode sub-vectors %
~\cite{MMWM,HeWS13}, but \emph{at the cost of higher quantization error}. Hashing hashes a sub-vector to one of the $K$ buckets. For example, \textsc{maddness} selects a 4-level balanced binary regression tree of the hashing function family, with each leaf as a hash bucket. A sub-vector is encoded by traversing the tree from the root, and moving to left or right child if the value of certain indices is above or below a threshold.    

This paper will evaluate both distance-based encoding and hashing-based encoding methods.

\vspace{-0.05in}
\subsection{PQ for AMM}
\label{sec:pqAMM}
\vspace{-0.05in}

PQ can be used for approximated matrix multiplication (AMM)~\cite{MMWM}. The essence is to approximate the matrix multiplication by the centroids' multiplication. 

To formulize it, for matrix multiplication $A\times B^T$, $a$ and $b$ are the rows of $A$ and $B$ respectively. The centroid codebooks for $A$ is $P$. For a layer of DNN, $A$ can be the input feature maps, and $B$ can be the weights (note that convolution can be computed as matrix multiplication too).  Since $B$ is constant, the multiplication of all the centroids and $B$ can be precomputed to construct a lookup table, as shown in Fig.~\ref{fig:product_quantization}(b). The table construction function for $b^c$ is shown in Eq.~\ref{eq:table}. 
\vspace{-0.05in}
\begin{equation}
\begin{split}
h^{c}(b^{c})=[P_0^c \cdot b^{c}, P_1^c \cdot b^{c}, \cdots, P_{K-1}^c \cdot b^{c}]
\end{split}
\label{eq:table}
\end{equation}
\vspace{-0.2in}

The matrix multiplication can then be approximated by looking up and aggregating the results of the nearest centroids in the precompute table, formulated in Eq.~\ref{eq:aggregation}. Here considers the $g^c(a^c)$ function with $onehot$ representation for $argmin$, i.e., the nearest centroid is marked as 1 and others as 0, $g^c(a^c)=onehot(\underset{k}{\mathrm{arg\ min}}\left \| a^{c}-P^{c}_{k} \right \|^2)=(0, ..., 0, 1, 0, ..., 0)$. 
\vspace{-0.15in}
\begin{equation}
a \cdot b = \sum_c {a^{c}}b^{c} \approx \sum_c {g^{c}(a^{c})}\cdot h^{c}(b^{c})
\label{eq:aggregation}
\end{equation}
\vspace{-0.25in}

\subsection{Motivation: Poor results of PQ for DNN}
\vspace{-0.05in}

Since DNN models are composed of MM, a direct thought is that can we replace MM in a DNN model by the PQ-based AMM. However, results show that directly applying it to a DNN model leads to poor accuracy.

\begin{figure}
    \centering
    \subfloat[Vanilla PQ-based AMM]{\includegraphics[width=0.23\textwidth]{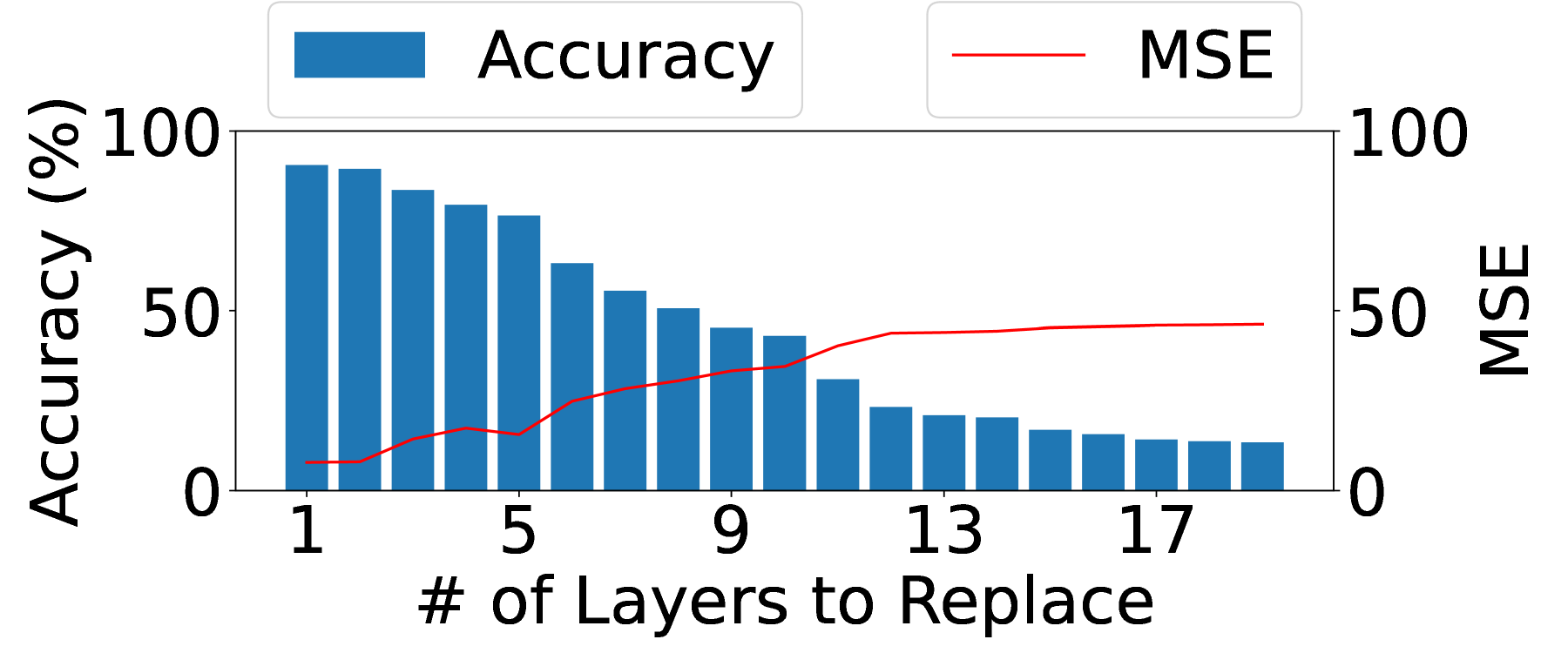}\label{fig:ammfordnn-kmeans}}
	\subfloat[\textsc{maddness} AMM]{\includegraphics[width=0.23\textwidth]{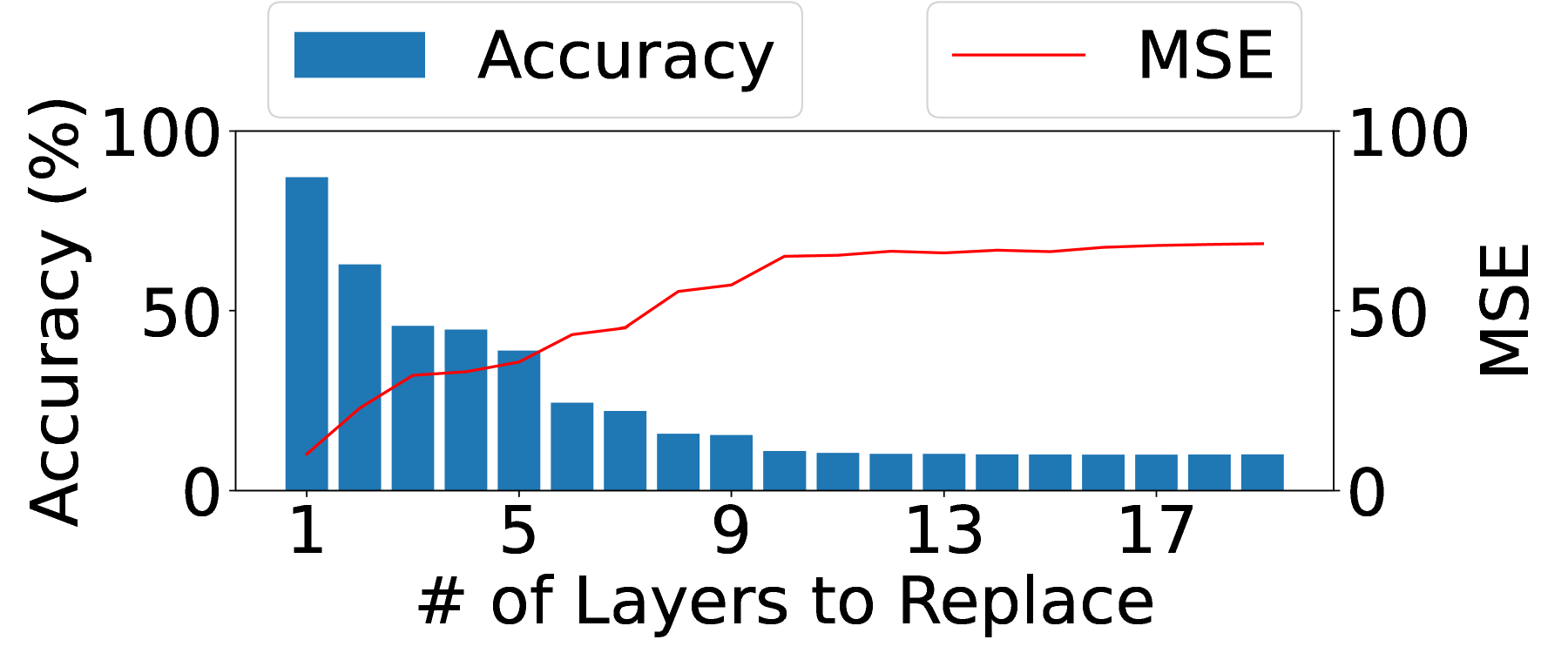}\label{fig:ammfordnn-hash}}
    \vspace{-0.15in}
    \caption{
    Model accuracy keeps decreasing because MSE keeps increasing, while more MMs (from the last to the first) are replaced by PQ-based AMM, for ResNet20 on CIFAR-10. \textsc{maddness} leads to more accuracy loss because using hashing for encoding.%
    }
    \vspace{-0.3in}
    \label{AMMforDNN}
\end{figure}

Fig.~\ref{AMMforDNN} shows the accuracy of using PQ-based AMM for ResNet-20 on CIFAR-10, as well as the Mean Square Error (MSE) of the replaced model and the original model. We replace the MM from the last to the first layer by PQ-based AMM. Fig.~\ref{fig:ammfordnn-kmeans} uses vanilla PQ with $k$-means for encoding, while Fig.~\ref{fig:ammfordnn-hash} uses \textsc{maddness} with hashing for encoding.

Results show the accuracy keeps dropping while more layers are replaced by AMM, because the error of the AMMs is accumulated. As expected, vanilla PQ shows better results than \textsc{maddness}, since $k$-means introduces smaller quantization error than hashing. \textsc{maddness} can maintain accuracy when only the last layer is replaced.   
This is consistent with the \textsc{maddness} paper, which only replaces the last layer, i.e., fully-connect, by AMM. However, if we replace the last two layers, the accuracy sharply drops by 30\%, and finally ends in 10\% accuracy. Vanilla PQ drops by 30\% when the last six layers are replaced, and also ends in 10\% accuracy. 

\textbf{Reason for poor accuracy}  We expose the key reason for the poor accuracy of PQ-based AMM is that \emph{the optimization goal of PQ and DNN learning is different}. As shown in Eq.~\ref{eq:pqgoal}, the goal of PQ is to minimize quantization error, i.e., learn the centroids to minimize the distance of each sub-vector and its nearest centroid.
On the other hand, the learning goal of DNN is to minimize the final loss function, through backpropagation to iteratively adjust the model parameters of each layer. Without considering the loss function, when more layers use AMM, the approximation error gets accumulated, as shown in Fig.~\ref{AMMforDNN}. %
To get better accuracy, it is necessary to learn the centroids by the DNN training process. 

However, the challenge is that PQ centroid learning and encoding functions shown in Eqs.~\ref{eq:pqgoal} and ~\ref{eq:encoding} are not differentiable, and cannot use backpropagation to calculate gradient.
This paper thus proposes the soft-PQ technique, which empowers the centroid learning for a DNN by backpropagation and gradient descent.

\vspace{-0.15in}
\section{Differentiable Centroid Learning for DNN}
\vspace{-0.05in}
As exposed above, the key factor to empower table lookup for DNNs is to learn the centroids of each layer through backpropagation to minimize the model loss. However, the main challenge is that the \verb|argmin| function in vanilla PQ, according to Eq.~\ref{eq:encoding}, is not differentiable. We therefore propose the differentiable centroid learning for DNN. The learning integrates three methods to adapt three levels of approximation. 

Firstly, we propose soft-PQ method, which leverages the continuous and differentiable approximation to \verb|argmax| function, i.e., \verb|softmax|~\cite{jang2016}, for backpropagation. We approximate the $onehot$ centroid result as the weighted sum of all centroid results (Sec.~\ref{sec:soft_encoding_function}). 
Prior work has used \verb|softmax| for differentiable PQ, but only for input embedding compression\cite{chen2020differentiable}. Our work is the first to enable soft-PQ for end-to-end DNNs.

Secondly, the use of \verb|softmax| introduces another level of approximation to using \verb|argmin|, which could also lead to reduced model accuracy.   %
We propose a learnable temperature method to address this challenge to adjust the approximation error of \verb|softmax| with \verb|argmin| for each layer. 
The temperature of each layer's \verb|softmax| can be learned through backpropagation (Sec.~\ref{sec:temperature}).

Thirdly, to reduce memory cost of the lookup tables, \sysname applies scalar quantization on the lookup tables. However, this also introduces a level of approximation. Similarly, we propose quantization-aware training to adapt this approximation during centroid learning (Sec.~\ref{scalar-quantized lut}).

\begin{figure}[!hbt]
  \centering
  \vspace{-0.15in}
  \includegraphics[width=0.75\linewidth]{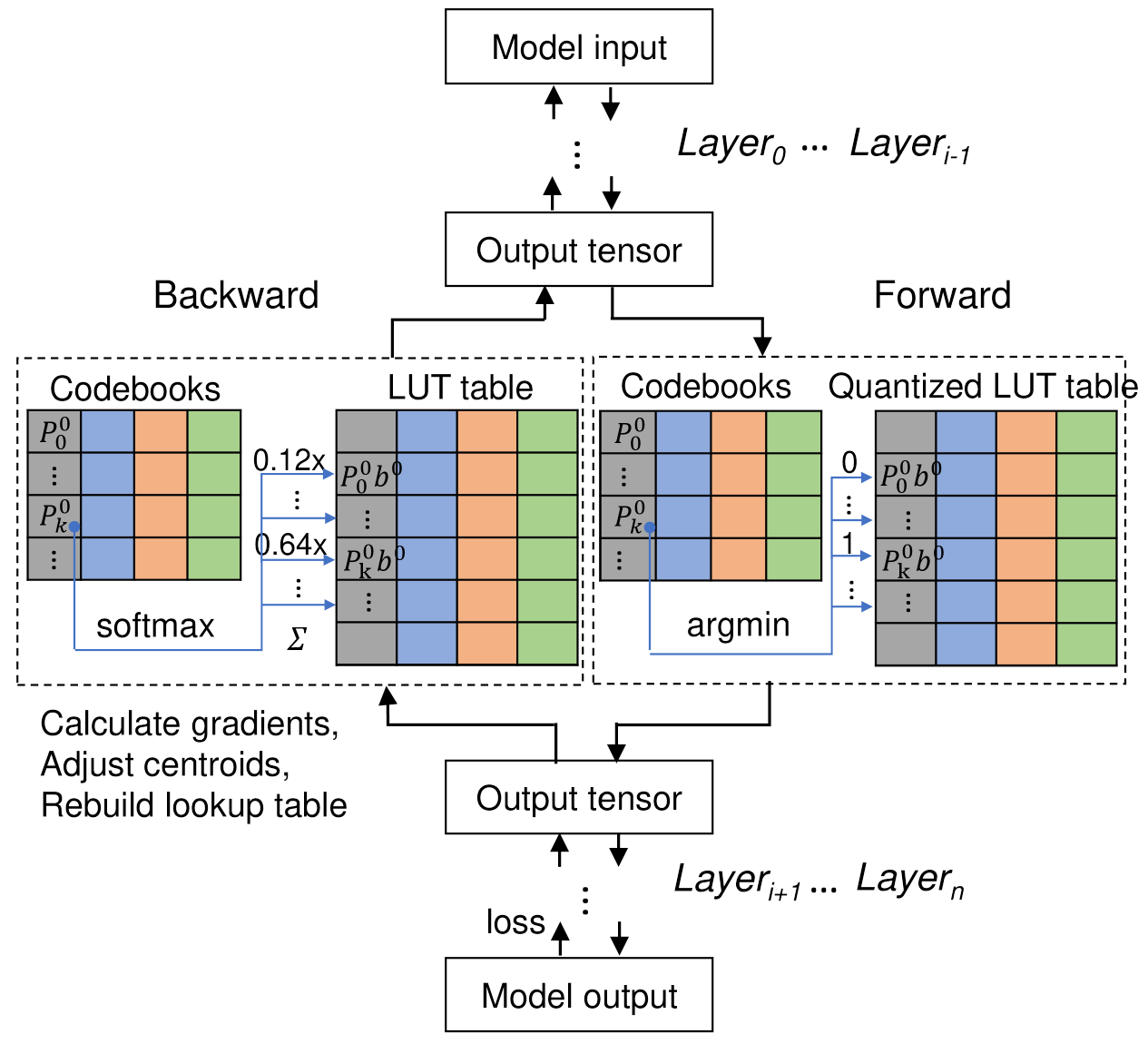}
  \vspace{-0.15in}
  \caption{
  \label{fig:training}
   Soft PQ learns centroids. Forward pass uses $\mathrm{argmin}$ as the encoding function, and backward uses $\mathrm{softmax}$. Each color marks a sub-vector dataset using one codebook. $P^c_k$ is the $k^{th}$ centroid for sub-vector $a^c$.
  }
  \vspace{-0.15in}
\end{figure}

\vspace{-0.12in}
\subsection{Backpropagation through soft-PQ}
\label{sec:soft_encoding_function}
\vspace{-0.05in}

As discussed in Sec.\ref{sec:pq}, vanilla PQ employs $k$-means to learn centroids from the dataset, and the encoding function $g^c(a^c)$ utilizes \verb|argmin| to encode a sub-vector as the nearest centroid, represented by a \verb|onehot| vector $(0,...,1,...,0)$ as shown in Fig.\ref{fig:training}. The sub-vector AMM result can be read directly from the lookup table by $g^c(a^c)\cdot h^c(a^c)$.

However, to apply PQ to the entire DNN model and minimize the model loss, the centroids for each layer must be learned from backpropagation and gradient descent. We utilize the smooth \verb|argmax| function, i.e., \verb|softmax|, as the encoding function for backpropagation, as shown in Eq.\ref{eq:softmax}. Here, $t$ represents the \verb|temperature| hyperparameter, which will be further discussed in Sec.~\ref{sec:temperature}.

\vspace{-0.18in}
\begin{equation}
\tilde{g}^{c}(a^{c})=\mathrm{softmax}(-\left \|a^{c}-P^{c}_{k}\right \|^2 /\ t)
\label{eq:softmax}
\end{equation}

For a codebook with $K$ centroids, the \verb|softmax| function takes a vector of $K$ distance results between the sub-vector $a^c$ and each centroid $P^c_k$ as input. 
It normalizes the input to a probability distribution that adds up to 1. According to the definition of \verb|softmax|, each probability is proportional to the exponent of the distance, i.e., $exp({-\left\| a^c-P^c_k \right\|^2}/{t})$. 
The closer the centroid is to the sub-vector, the higher the probability will be. 
The encoding for a sub-vector is transformed from a deterministic \verb|onehot| vector into a probability vector. 
For example, as illustrated in Fig.~\ref{fig:training}, the output of the \verb|softmax| encoding function is $(0.12,...,0.64,...)$, where 0.64 is the probability of the nearest centroid. The sub-vector AMM result is then obtained as the dot product of the probability vector and the lookup table entries.

\textbf{Soft-PQ centroid learning:} Using \verb|softmax|, the centroid learning process of our soft-PQ for the entire model is illustrated in Fig.\ref{fig:training}. 
During the forward pass, the onehot \verb|argmin| function is utilized to calculate the model output and loss, as model inference will also use \verb|argmin| for simplicity. 
The backward pass utilizes \verb|softmax| as the encoding function to calculate gradients, adjust centroids via gradient descent, and rebuild lookup tables with the updated centroids for the next training iteration. 
Based on Eq.\ref{eq:aggregation}, the sub-vector AMM in soft-PQ is formulated as Eq.~\ref{eq:training}.
\vspace{-0.05in}
\begin{equation}
a^cb^c=\tilde{g}^c(a^c)\cdot h^c(b^c)-\mathrm{sg}(\tilde{g}^c(a^c)\cdot h^c(b^c)-g^c(a^c)\cdot h^c(b^c))
\label{eq:training}
\end{equation}
\vspace{-0.2in}

Here, $sg$ represents the \emph{stop gradient} operator. It serves as an identity function during the forward pass to enable the use of $g^c(a^c)$ encoding in \verb|argmin|. During the backward pass, it drops gradients inside it to enable $\tilde{g}^c(a^c)$ to generate gradients via \verb|softmax|.

The initial value is critical for learning convergence and accuracy. Therefore, we use the centroids learned by $k$-means from vanilla PQ to initialize the centroids and lookup tables.

\vspace{-0.1in}
\subsection{Learned temperature}
\label{sec:temperature}

\begin{figure}[!hbt]
    \centering
    \vspace{-0.25in}
    \subfloat[]{
        \includegraphics[width=0.25\columnwidth]{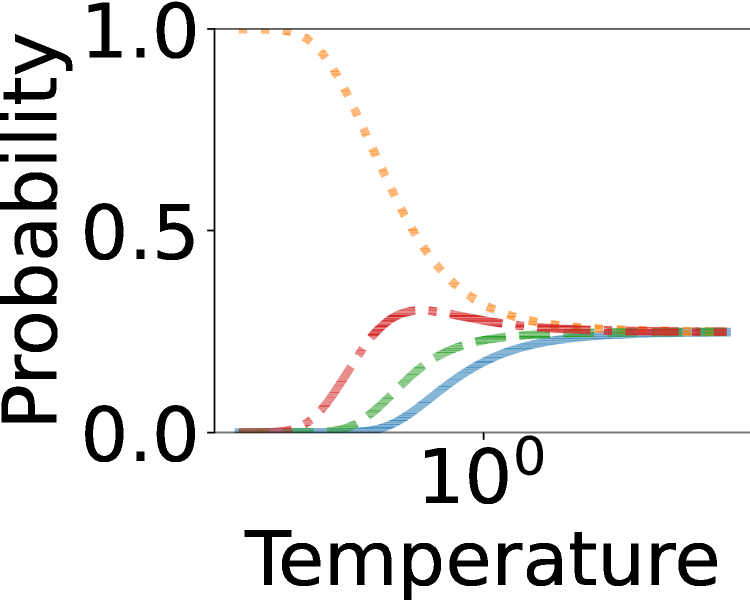}
        \label{fig:softmax}
    }
    \hspace{0.25in}
    \subfloat[]{
        \includegraphics[width=0.26\columnwidth]{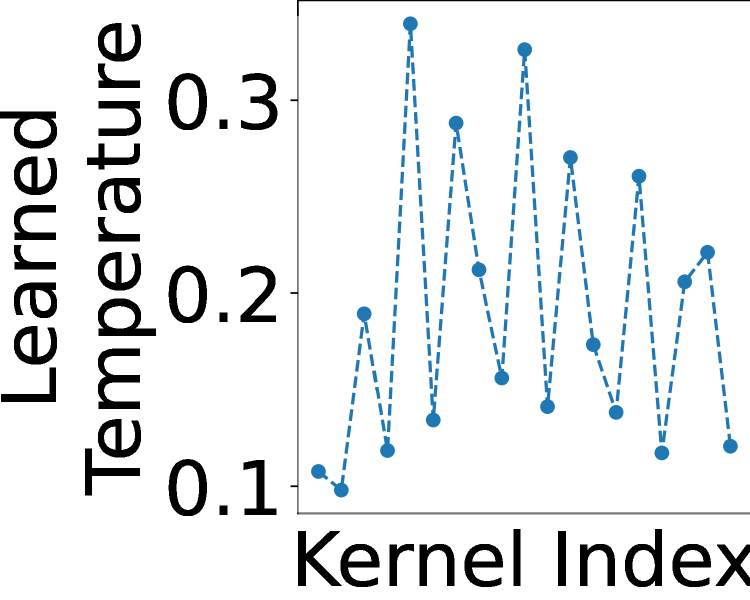}
        \label{tab:learnt}
    }
  \vspace{-0.15in}
  \caption{
  (a) The output probability distribution of $\mathrm{softmax}(\frac{x}{t})$ at different temperature, for four centroids as an example. $t\rightarrow\infty$ approaches uniform distribution, and $t\rightarrow 0$ approaches $\mathrm{argmax}$. (b) The learned $t$ for each layer of ResNet18.
  }
  \vspace{-0.3in}
\end{figure}

The \verb|temperature| hyperparameter $t$ of \verb|softmax|~\cite{hinton2015} controls the approximation error of \verb|softmax| to \verb|argmax|. Shown in Fig.~\ref{fig:softmax}, as $softmax(x)_i=\frac{exp(x_i/t)}{\sum_{k=1}^K exp(x_k/t)}$,  when $t\rightarrow\infty$, $softmax(x)_i\rightarrow\frac{1}{K}$, i.e., the output probability distribution approaches uniform distribution. When $t\rightarrow0$, $softmax(x)\rightarrow onehot(argmax(x))$, i.e., the probability of the largest $x$ approaching 1 and the others is 0. 

Therefore, there is a tradeoff between small and large temperature. For small temperature, \verb|softmax| is close to \verb|onehot| \verb|argmax|, but the training is difficult since the variance of gradients is large. For larger temperature, the approximation error is increased, but the variance of gradients is smaller. 

Works before normally set $t$ as a fixed value (mostly 1), or anneal it from a large number to a small one during training~\cite{hinton2015,jang2016}, but never analyze how to set it reasonably. This is because currently for a DNN model, \verb|softmax| is only used by the output layer to produce class probability, or used by the input layer to produce symbol embedding. The approximation error barely impacts model accuracy. However, our soft-PQ employs \verb|softmax| in every layer. The accumulated error can decrease accuracy without proper $t$ settings.          

We thus propose to learn the temperature for each layer, also during backpropagation while centroid learning. Fig.~\ref{tab:learnt} shows the learned $t$ for each layer for ResNet18. The value for each layer is different, and thus not practical to tune by hand.
According to the CIFAR10 accuracy experiments,
training with the learned temperature technique only spends $\frac{1}{10}$ iterations of training with temperature setting to 1 to achieve 85\% accuracy.
Detailed results will show in Sec.~\ref{sec:eval}.

\subsection{Scalar quantized lookup table}
\label{scalar-quantized lut}
\vspace{-0.05in}

Lookup tables are the main disk and memory cost. We reduce the table size by scalar quantization (e.g., FP32 to INT8). We leverage the classic range-based linear quantization. The formula is $r=s(q-z)$~\cite{jacob2018quantization}, in which $r$ is the real value, $s$ is the scaling factor, $q$ is the quantized value, and $z$ is the zero point. 
We use symmetric quantization, so $z$ is forced to be 0, and the quantized range is $[-2^{n-1}, 2^{n-1}-1]$. The scaling factor $s$ is calculated as the max absolute value in the table divided by half of the range, i.e., $s=\frac{\text{max(|value|)}}{2^{n-1}-1}$.

Quantized lookup tables introduce another level of approximation. Similar to \verb|temperature|, we thus quantize the tables during centroid learning, to minimize the loss function. Inspired by Jacob \etal~\cite{jacob2018quantization},  the backpropagation uses lookup tables in real values, so that they can be adjusted in small
amounts. The forward pass uses quantized lookup tables as in the inference to calculate the loss. Results show that by this learning method, the quantized lookup table has little impact on the model accuracy. %

\vspace{-0.15in}
\section{cost reduction of \sysname}
\label{cost analysis}
Before introducing the inference design of \sysname, this section analyzes the theoretical FLOPs and model size reduction by \sysname.  
The two primary factors of \sysname are the number of centroids and the length of sub-vectors, representing a tradeoff between cost and accuracy. 
We conduct the analysis %
with these two factors.

According to the output formula Eq.~\ref{eq:aggregation} of a PQ-based AMM in \sysname, the main cost is from the encoding function $g^c(a^c)$, which calculates the Euclidean distance of the sub-vector with each centroid. After that, the cost is from table lookup with the encoding result (i.e., index of the closest centroid), and the result aggregation of sub-vectors. For file size, the major cost is from the lookup tables, which saves the dot product result of each centroid and the according sub-vectors in the weight matrix. The size of codebooks is relatively small, since the sub-vectors on the same column share one codebook.  

Therefore, we analyze the FLOPs of encoding, table lookup and aggregation, as well as the size of lookup tables as the cost of a \sysname AMM, to compare it with normal MM in Table~\ref{tab:syscost}. Since convolution can be transformed to MM by im2col, its cost also follows these formulas. For a convolution, $M$ is the number of output channels, $D$ is the number of input channels $\times$ filter size$^2$, and $N$ is height $\times$ width.    

The number of centroids $K$ and the sub-vector length $V$ are two hyperparameters of \sysname. They are tradeoffs between accuracy and cost (refer to Sec.~\ref{sec:ablation}).  
The more centroids $K$ and shorter sub-vector $V$ may lead to higher accuracy, but will increase the cost of \sysname. Table~\ref{tab:flops} shows the calculated GFLOPs and model size by the formulas in Table~\ref{tab:syscost} for different models, using typical $(K,V)$ settings in \sysname. These typical settings can achieve comparable accuracy with the original model, and also align with the SIMD width for high performance. Similar to other hyperparameters in DNN training, $K$ and $V$ can be set by grid search, evolutionary search, or other popular searching methods.%

It is clear that \sysname can achieve both computation and model size saving. The FLOPs saving is because $K$ is normally smaller than $M$. For example, the number of output channels i.e., $M$, for ResNet50 is normally 128, 256, or 512, so the FLOPs can be reduced by $4\times$ when $K=8$. For ResNet20, the number of output channel is 16, 32, and 64, so the FLOPs is reduced by $2\times$ when $K=8$. %

\begin{table}[!hbt]
 \centering
  \vspace{-0.1in}
  \begin{tabular}{|c|c|c|}
  \hline
  \multicolumn{3}{|l|}{$A\in \mathbb{R}^{N\times D}$: input matrix} \\
  \multicolumn{3}{|l|}{$B\in \mathbb{R}^{D\times M}$: weight matrix} \\
  \multicolumn{3}{|l|}{$V$: the length of a sub-vector $a^c$} \\
  \multicolumn{3}{|l|}{$K$: the number of centroids in a codebook for $a^c$} \\
  \hline
  & \textbf{Ours} & \textbf{MM} \\
  \hline
  \textbf{FLOPs} & $N \cdot D \cdot K + N \cdot M \cdot D / V $ & $N \cdot D \cdot M$\\
  \hline
  \textbf{Size} & $4\cdot D\cdot K + D \cdot M \cdot K /\, V$ & $4 \cdot D \cdot M$ \\
  \hline
  \end{tabular}
    \caption{FLOPs and disk size of a \sysname AMM compared to MM (FP32).}
    \vspace{-0.4in}
  \label{tab:syscost}
\end{table}
\begin{table}[!hbt]
\centering
\begin{tabular}{|c|ccc|} 
\hline
\multirow{2}{*}{Model} & \multicolumn{3}{c|}{GFLOPs}     \\ 
\cline{2-4}
                       & original & (8, 9)   & (16, 9)   \\ 
\hline
ResNet18 (CIFAR10)     & 0.555    & 0.098    & 0.132     \\
SENet18 (CIFAR10)      & 0.556    & 0.098    & 0.131     \\
VGG11 (CIFAR10)        & 0.606    & 0.085    & 0.102     \\
ResNet18               & 1.814    & 0.412    & 0.515     \\
SENet18                & 1.814    & 0.412    & 0.515     \\
VGG11                  & 7.609    & 1.132    & 1.334     \\ 
\hline
                       & original & (16, 32) & (16, 16)  \\ 
\hline
BERT                   & 2.759    & 0.169    & 0.254     \\
\hline
\end{tabular}
\centering
\begin{tabular}{|c|ccc|} 
\hline
\multirow{2}{*}{Model} & \multicolumn{3}{c|}{Disk Size (MB)}  \\ 
\cline{2-4}
                       & original & (8, 9)   & (16, 9)        \\ 
\hline
ResNet18 (CIFAR10)     & 42.59    & 12.55    & 23.13          \\
SENet18 (CIFAR10)      & 42.92    & 10.94    & 21.53          \\
VGG11 (CIFAR10)        & 35.18    & 8.45     & 16.88          \\
ResNet18               & 44.55    & 12.57    & 23.16          \\
SENet18                & 44.88    & 12.91    & 23.49          \\
VGG11                  & 37.12    & 10.39    & 18.82          \\ 
\hline
                       & original & (16, 32) & (16, 16)       \\ 
\hline
BERT                   & 162.26   & 23.05    & 43.30          \\
\hline
\end{tabular}
\caption{
\label{tab:flops}
Theoretical GFLOPs and model size of typical ($K$, $V$), calculated by formulas in Table~\ref{tab:syscost}. 
}
\vspace{-0.45in}
\end{table}

\section{Table lookup Inference Design}
\label{sec:inference_design}

\begin{figure}[!htb]
    \centering
    \vspace{-0.12in}
    \includegraphics[width=0.8\linewidth]{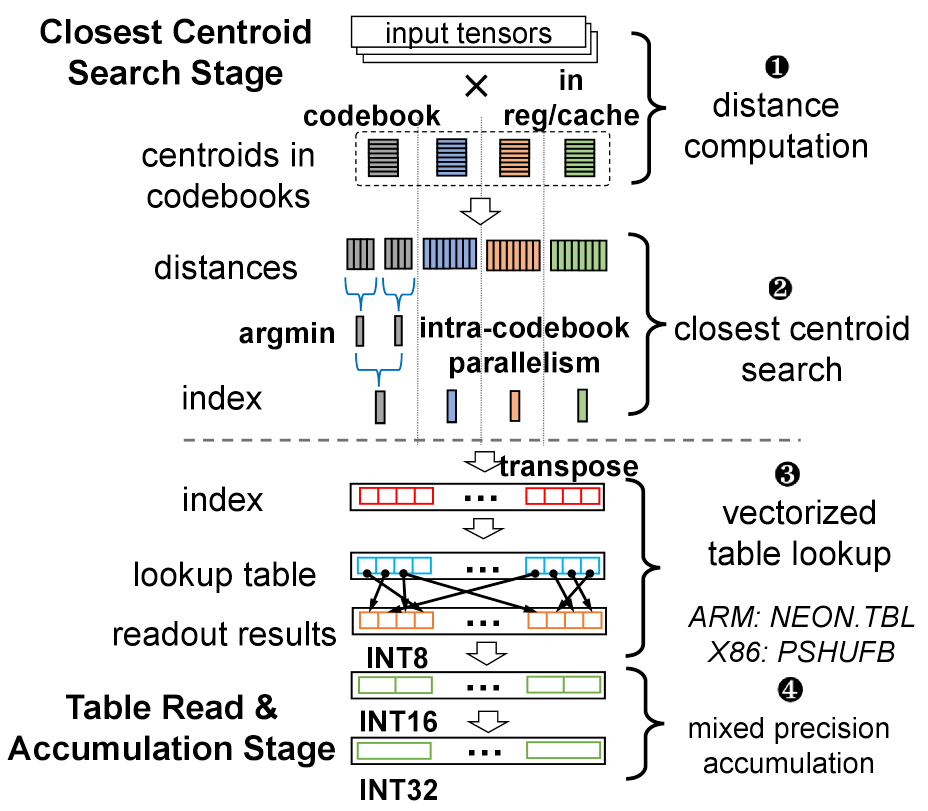}
    \vspace{-0.15in}
    \caption{Table Lookup Inference Design}
    \vspace{-0.3in}
    \label{fig:operator_design}
\end{figure}

In this section, we present the design and optimization of the inference system that supports \sysname. 
Fig.~\ref{fig:operator_design} depicts the \sysname model inference architecture, comprising the Closest Centroid Search stage and the Table Read and Accumulation stage. 
In the Closest Centroid Search stage, \sysname first computes the distance between input tensors and centroids and searches the nearest centroids for each input tensor.
In the Table Read and Accumulation stage, \sysname reads the pre-computed results from the lookup table and accumulate results.
\sysname inference design fully takes advantage of the features of current hardware architectures to optimize the design and implementation of Table Lookup. 
These optimizations improve inference by utilizing memory hierarchy, utilizing shuffle instructions, and utilizing mixed-precision accumulation instructions.

\vspace{-0.05in}
\subsection{Closest centroid search}
\vspace{-0.05in}
The centroid search takes the most computation costs in \sysname, and efficient centroid search is critical for performance efficiency compared to conventional computational methods. 
The \textbf{Closest Centroid Search} stage first calculates the distance between input tensors and centroids. 
We can represent this calculation as matrix multiplications of input tensors and codebook matrices. 
Then, it searches for the nearest centroids for each input sub-vector.

However, the design of centroid search in \sysname presents challenges that involve leveraging the features of existing hardware architectures. 
Firstly, \sysname distance computation is irregular-shaped (tall-and-skinny) and difficult to optimize using BLAS libraries. 
For example, the XNNPACK library achieves only $23.0$ GFLOP/s on Pixel 6 for the second layer of LUT-NN based ResNet18, which accounts for only $25.7\%$ of peak performance. 
The height of the input tensor ($N$) is usually much larger than the length of the sub-vector ($N \gg V$) and the number of centroids ($N \gg K$) on each codebook. 
Therefore, the operation intensity of the distance calculation can be approximated by $\frac{2NVK}{NV + KV + NK} \approx \frac{2}{1/K + 1/V} \text{FLOP/Byte}$~\cite{wang2021asymo}.
Since the length of the sub-vector and the number of centroids are small in \sysname, the operation intensity $\frac{2}{1/K + 1/V}$ is also small. 
The distance computation becomes a memory-intensive matrix multiplication.

Therefore, we focus primarily on optimizing memory access for centroid distance computations (\ding{182} in Fig.~\ref{fig:operator_design}) to leverage memory hierarchy in current hardware.
To optimize memory access for centroid distance computations (\ding{182} in Fig.~\ref{fig:operator_design}), we reduce memory access overhead by keeping frequently accessed data in registers and caches as much as possible.
So we design a \emph{centroid-stationary computation} scheme to reside centroid matrices in registers and reorder centroid matrix loads in the inner loop to keep them in the cache as long as possible. 
The centroid-stationary computation keeps $K$ centroids in the cache for each codebook, which only requires reading these centroids once from DRAM. 
Consequently, it only requires reading an $N \cdot V$ input tensor from DRAM once, reducing memory bandwidth costs and improving performance.

Second, after computing centroid distances, the Closest Centroid Search stage must identify the nearest centroid for each input sub-vector and generate the centroid index (\ding{183} in Fig.~\ref{fig:operator_design}). 
It can be represented by an \verb|argmin| function, which finds an index with the shortest distance. 
However, searching for the nearest centroid for each input sub-vector is a data-dependent operation. 
To find the closest one, we must compare each distance sequentially, which is RAW (Read After Write) dependent and hard to be parallelized on CPUs.

We propose\emph{intra-codebook parallelism} to optimize the Closest Centroid Search stage. 
{Intra-codebook parallelism} searches the nearest centroid for the input sub-vector on a codebook in parallel. 
We slice a codebook into multiple sub-codebooks and compare each distance between sub-vector and centroids in each sub-codebook. 
Then, we interleave each sub-codebook's execution, and avoid data dependency in closest centroid search.
We merge the compared distances by reduction and find the index corresponding to the closest centroid. 
It leverages instruction-level parallelism to improve hardware utilization and performance.

\vspace{-0.05in}
\subsection{Table read and accumulation} 
\vspace{-0.05in}
\sysname obtains the indices of the nearest centroids for the input sub-vectors after closest centroid search, and leverages \textbf{Table Read and Accumulation} stage to compute the final results. 
This stage first reads the pre-computed results from the corresponding lookup table through indices (\ding{184} in Fig.~\ref{fig:operator_design}) and completes the computation by accumulation operation (\ding{185} in Fig.~\ref{fig:operator_design}).
For example, convolution operators directly read out filter's outputs from lookup tables and accumulate each input channel's result for output channels. 

However, Table Read and Accumulation stage introduces additional overhead in inference. 
We enhance inference efficiency by skillfully utilizing widely available and supported instructions in commercial hardware. 
First, table read is difficult to parallelize and introduces additional indirect memory accesses, which exaggerate the memory overhead on lookup tables. 
Since we quantize lookup table into INT8 in Sec.~\ref{scalar-quantized lut}, we leverage widely supported SIMD \emph{shuffle} instructions (x86: \verb|pshfb| and ARM: \verb|tbl|) to achieve parallel and efficient table read. 
We demonstrate the implementation of table read using shuffle instructions in Fig.~\ref{fig:operator_design}.  
The shuffle instruction \cite{blalock2017bolt} permutes each byte of a vector based on an index vector and stores the shuffled bytes in the result vector register in each clock cycle. 
On 128-bit wide SIMD, a vectorized table read instruction handles 16 sub-vector lookups ($128 / 8 = 16$) on $16$ results ($128 / 8 = 16$) simultaneously, greatly simplifying table read and reducing overheads.

Second, the accumulation still has comparable computation costs to the entire process.
For example, when a codebook handles $N$ index lookups on a $K \cdot M$ lookup table, it costs $N \cdot M$ table reads ($K = 16$) and $N \cdot M$ accumulation adds. 
Therefore, the vectorized lookup table leaves accumulation operations as the performance bottleneck of table reads. 
The number of parallel processing units within a SIMD instruction is called SIMD lanes. 
We typically set the number of centroids to 16 (K=16) to maximize the utilization of all SIMD instruction lanes.
Since a higher number of SIMD lanes leads to higher throughput on the same width of SIMD instruction (e.g., the vectorized INT16 add instruction has the half throughput of INT8 on a 128-bit SIMD), 
we maximize accumulation throughput by mixed precision accumulation. It first accumulates results in INT16 to utilize more SIMD lanes and then gathers INT16 to INT32 to avoid overflow.

\add{
\vspace{-0.1in}
\subsection{Utilization of memory hierarchy}
\vspace{-0.05in}

As \sysname is composed of many memory accesses, we explain more details for its memory hierachy usage. 

The frequent-accessed codebook (for centroids) is kept in the inner loop during distance calculation, so that the hardware cache can leaverage the locality to keep the codebook in cache.  
The size of a codebook for one layer is normally small, in KB level (calculated by sub-vector length $\times$ number of centroids $\times$ number of codebooks $\times$ number of bytes).  For example, the largest codebook in our tested models is 288\,KB. It can be held in L2 cache on many mobile devices.

The lookup table (for precomputed results) is larger, in MB level. It can be held in L3 cache. Even if it is larger than L3 cache, the memory access overhead could still be hidden in our implementation. LUT-NN uses Shuffle instructions to parallelize table lookup, turning the pattern of reading index numbers and lookup tables from random to sequential. The memory access is thus predictable and easy to prefetch. Furthermore, we interleave computation and memory access instructions within the kernel, which can overlap computation and memory access time.
}

\begin{table*}[tbh!]
\centering

\begin{tabular}{|c|c|c|c|c|c|c|c|c|}
\hline
Dataset &
  Model &
  \begin{tabular}[c]{@{}c@{}}Centroid \\ Learning Rate\end{tabular} &
  \begin{tabular}[c]{@{}c@{}}Temperature\\ Learning Rate\end{tabular} &
  \begin{tabular}[c]{@{}c@{}}Weight\\ Decay\end{tabular} &
  \begin{tabular}[c]{@{}c@{}}Batch\\ Size\end{tabular} &
  \# of Epochs &
  Optimizer &
  \begin{tabular}[c]{@{}c@{}}LR\\ Scheduler\end{tabular} \\ \hline
CIFAR10 &
  \multirow{6}{*}{\begin{tabular}[c]{@{}c@{}}ResNet18\\ SENet18\\ VGG11\end{tabular}} &
  \multirow{2}{*}{$1\times 10^{-3}$} &
  \multirow{7}{*}{$1\times 10^{-1}$} &
  \multirow{6}{*}{0} &
  \multirow{6}{*}{256} &
  \multirow{3}{*}{200} &
  \multirow{6}{*}{Adam} &
  \multirow{6}{*}{\begin{tabular}[c]{@{}c@{}}Cosine\\ Annealing\end{tabular}} \\ \cline{1-1}
GRSRB           &      &                                    &  &                   &    &     &       &          \\ \cline{1-1} \cline{3-3}
Speech Commands &      & \multirow{2}{*}{$1\times 10^{-4}$} &  &                   &    &     &       &          \\ \cline{1-1} \cline{7-7}
\add{UTKFace}         &      &                                    &  &                   &    & 150 &       &          \\ \cline{1-1} \cline{3-3} \cline{7-7}
SVHN            &      & \multirow{2}{*}{$1\times 10^{-3}$} &  &                   &    & 200 &       &          \\ \cline{1-1} \cline{7-7}
ImageNet        &      &                                    &  &                   &    & 150 &       &          \\ \cline{1-3} \cline{5-9} 
GLUE            & BERT & $\{5,4,3,2\} \times 10^{-5}$       &  & $1\times 10^{-2}$ & 32 & 3   & AdamW & Constant \\ \hline
\end{tabular}

\caption{
\label{tab:exper_settings}
Soft-PQ training settings for centroid and temperature learning. 
We choose the best fine-tuning learning rate among \{$5\times 10^{-5}$, $4\times 10^{-5}$, $3\times 10^{-5}$, $2\times 10^{-5}$\} based on the accuracy of the validation set for BERT, following \cite{devlin2018bert}.
}
\vspace{-0.32in}
\end{table*}
\begin{table*}[tbh!]
\centering

\begin{tabular}{|c|ccc|ccc|ccc|}
\hline
Model & \multicolumn{3}{c|}{ResNet18} & \multicolumn{3}{c|}{SENet18} & \multicolumn{3}{c|}{VGG11} \\ \hline
Dataset & \multicolumn{1}{c|}{LUT-NN} & \multicolumn{1}{c|}{MADDNESS} & baseline & \multicolumn{1}{c|}{LUT-NN} & \multicolumn{1}{c|}{MADDNESS} & baseline & \multicolumn{1}{c|}{LUT-NN} & \multicolumn{1}{c|}{MADDNESS} & baseline \\ \hline
CIFAR10 & \multicolumn{1}{c|}{94.40} & \multicolumn{1}{c|}{10.01} & 95.26 & \multicolumn{1}{c|}{94.48} & \multicolumn{1}{c|}{10.65} & 95.47 & \multicolumn{1}{c|}{93.89} & \multicolumn{1}{c|}{22.87} & 95.04 \\ \hline
GTSRB & \multicolumn{1}{c|}{98.73} & \multicolumn{1}{c|}{4.53} & 98.80 & \multicolumn{1}{c|}{98.36} & \multicolumn{1}{c|}{5.68} & 98.84 & \multicolumn{1}{c|}{98.55} & \multicolumn{1}{c|}{5.70} & 99.22 \\ \hline
\begin{tabular}[c]{@{}c@{}}Speech\\ Commands\end{tabular} & \multicolumn{1}{c|}{93.70} & \multicolumn{1}{c|}{1.49} & 91.72 & \multicolumn{1}{c|}{93.04} & \multicolumn{1}{c|}{1.49} & 94.36 & \multicolumn{1}{c|}{93.38} & \multicolumn{1}{c|}{1.49} & 93.11 \\ \hline
SVHN & \multicolumn{1}{c|}{96.00} & \multicolumn{1}{c|}{20.68} & 96.67 & \multicolumn{1}{c|}{96.22} & \multicolumn{1}{c|}{20.12} & 96.60 & \multicolumn{1}{c|}{96.23} & \multicolumn{1}{c|}{29.97} & 96.62 \\ \hline
\add{UTKFace} & \multicolumn{1}{c|}{4.91} & \multicolumn{1}{c|}{10.51} & 5.57 & \multicolumn{1}{c|}{4.74} & \multicolumn{1}{c|}{11.02} & 5.46 & \multicolumn{1}{c|}{5.69} & \multicolumn{1}{c|}{24.57} & 5.85 \\ \hline
ImageNet & \multicolumn{1}{c|}{67.38} & \multicolumn{1}{c|}{0.10} & 69.76 & \multicolumn{1}{c|}{68.21} & \multicolumn{1}{c|}{0.17} & 70.63 & \multicolumn{1}{c|}{68.04} & \multicolumn{1}{c|}{0.16} & 68.33 \\ \hline
\end{tabular}

\caption{
\sysname achieves comparable accuracy 
with the original models, significantly outperforming \textsc{Maddness}. \add{UTKFace is the lower the better. Others are the higher the better.} %
}
\label{tab:cnn_accuracy}
\vspace{-0.32in}
\end{table*}
\vspace{-0.15in}

\begin{table*}[htb!]
\centering
\begin{tabular}{|c|c|c|cc|c|}
\hline
\multirow{2}{*}{Dataset  Task} & \begin{tabular}[c]{@{}c@{}}Single\\ Sentence\end{tabular} & \begin{tabular}[c]{@{}c@{}}Similarity and \\ Paraphrase\end{tabular} & \multicolumn{2}{c|}{\begin{tabular}[c]{@{}c@{}}Natural Language \\ Inference\end{tabular}} &  \\ \cline{2-6} 
 & SST-2 & QQP & \multicolumn{1}{c|}{QNLI} & RTE & Average \\ \hline
Training Dataset Size & 67k & 364k & \multicolumn{1}{c|}{105k} & 2.5k &  \\ \hline
Test Dataset Size & 1.8k & 391k & \multicolumn{1}{c|}{5.4k} & 3k &  \\ \hline
BERT base (\%) & 93.5 & 71.2 & \multicolumn{1}{c|}{90.5} & 66.4 & 80.4 \\ \hline
LUT-NN (\%) & 92.4 & 69.6 & \multicolumn{1}{c|}{87.4} & 64.7 & 78.5 \\ \hline
\end{tabular}
\caption{
\sysname achieves comparable accuracy to BERT-base on NLP tasks selected from GLUE. 
\label{tab:bert_accuracy}
}
\vspace{-0.4in}
\end{table*}

\section{Evaluation}
\label{sec:eval}

\subsection{Experiment methodologies and settings}
\label{sec:exper_method} 

\textbf{Dataset and Models} 
Our experiment tests {\sysname}'s effectiveness on image recognition, speech recognition, and NLP tasks. 
For image recognition and speech recognition, we evaluate VGG, ResNet, and SENet family models on different datasets, including CIFAR-10~\cite{cifar}, GTSRB (German Traffic Sign Recognition Benchmark)~\cite{GTSRB}, Google Speech Command~\cite{speech_command}, SVHN (Street View House Numbers)~\cite{SVHN}, \add{UTKFace~\cite{zhifei2017cvpr}}, and the large-scale complex ImageNet~\cite{imagenet} dataset. 
The CIFAR-10 dataset consists of $60K$ images in $10$ classes, with $50K$ for training and $10K$ for validation. 
The GTSRB dataset has $43$ classes of traffic signs, with $39K$ training images and $12K$ test images. 
Google Speech Command is an audio dataset containing $65K$ recordings of $30$ words, designed for keyword spotting on edge devices. 
SVHN is a dataset including $600K$ real-world images of house numbers for $10$ classes.
\add{UTKFace dataset consists of $20K$ face images labeled by age, gender, and ethnicity. We use the age prediction task to test the regression ability of \sysname. We will report Mean Average Error (MAE) for this task.}
ImageNet dataset has $1.28M$ images for training and $50K$ for validation. 

On the large ImageNet dataset, ResNet18 and SENet18 models use 7x7 convolution for the first layer and 3x3 max-pooling, while other datasets use 3x3 convolution for the first layer and no max-pooling.
The final three dense layers of the original VGG models are replaced by an average pooling layer and a dense layer to reduce model size for mobile deployment, following the general practice for VGG deployment~\cite{wang2021asymo}. Except for ImageNet, the first max pooling layer is removed for VGG. In evaluation results, the models for ImageNet use the model names, and the models for other datasets are marked as "Model (CIFAR10)" to differentiate.  

To evaluate {\sysname} for NLP tasks, we used BERT models and the tasks from GLUE benchmark~\cite{glue}.

\noindent\textbf{Soft-PQ training setting}
Table~\ref{tab:exper_settings} lists the detailed training hyperparameter settings. 
Except for the initial learning rate, we followed standard practices for all other training procedures~\cite{he2016deep,devlin2018bert}. 
As the learned temperature requires a larger learning rate to converge quickly to a sub-optimal solution, we use different learning rates for the centroid and temperature learning. 
Prior to soft-PQ training, we initialize centroids using $k$-means clustering. 
Specifically, we apply the original model inference to a randomly sampled sub-dataset (consisting of 1024 training samples) and collect each operator's inputs. 
We then utilize the $k$-means algorithm to cluster each operator's inputs and obtain the initial centroids.

\noindent\textbf{\sysname system settings} In order to balance accuracy and cost, we can specify various settings for \sysname, including the number of centroids in a codebook ($K$), the length of the sub-vector ($V$) (as shown in Table~\ref{tab:syscost}), and the operators to be replaced by table lookup.  
The default setting for $K$ and $V$ is to align with both the feature size and the length of SIMD instructions. 
Specifically, we set $(K, V) = (16, 9)$ for $3 \times 3$ convolution,  $(K, V) = (16, 4)$ for $1 \times 1$ convolution, %
We will also evaluate other settings in Sec.~\ref{sec:ablation} ablation study. 

For operators to replace, we replace all convolution operators for CNN models by table lookup except the first one. This is because as explained by Zhou \etal~\cite{ZhouNZWWZ16},  the layer interacted with the model input is very sensitive to accuracy. We also observe this phenomenon. Replacing the first layer by table-lookup can lead to obvious accuracy drop. %
This issue is even more serious for BERT. Replacing the first two layers results in 80\% accuracy loss (refer to Fig. \ref{fig:bert_num_layers_to_replace}).  
Therefore, we do not replace these layers by table lookup. %
For BERT, other than explicitly stated, we replace the fully connected operators of the last 6 layers. The results with different replaced layers for BERT will be shown in Section~\ref{sec:ablation} ablation study.

\noindent\textbf{Evaluation platforms} 
\sysname is implemented on ARM Neon and Intel SIMD ISA (Instruction Set Architecture), and can run in a single or multi threads. %
We evaluate \sysname on two mobile devices Google Pixel 4 and 6, which are equipped with Cortex-A76 ($2.42$ GHz) and Cortex-X1 ($2.80$ GHz), respectively. 
In addition, we evaluate \sysname on a desktop CPU (Intel Core i7-4790) running at $3.60$ GHz and a server CPU (Xeon Silver 4210) running at $2.20$ GHz.

\noindent\textbf{Comparison baselines} The accuracy baselines are (1) original models and (2) \textsc{Maddness}~\cite{MMWM}, respectively. The inference performance baselines are ONNX Runtime (ORT) v1.12.1~\cite{onnx} and %
TVM v0.9.0~\cite{chen2018tvm}, as the SOTA hand-written performance, and the auto-tuned performance upper bound for linear computation opeartors. We tune the TVM baselines using AutoScheduler and AutoTVM in $1500$ iterations.

\vspace{-0.2in}
\subsection{Accuracy, latency, memory and power}
\label{sec:e2e_evaluation}
\vspace{-0.05in}
\textbf{Accuracy}
Table~\ref{tab:cnn_accuracy} and~\ref{tab:bert_accuracy} summarize the accuracy comparison. Remarkably, to compare with directly applying \textsc{maddness}, \sysname can improve accuracy by 66\% to 92\%, achieving similar-level of accuracy with the original models. \sysname behaves particularly well on GTSRB ($\leq0.67\%$ difference), SVHN ($\leq0.67\%$ difference), CIFAR-10 ($\leq1.15\%$ difference), Speech Command ($\leq1.32\%$ difference) \add{, and UTKFace (LUT-NN is even better)}. For the Speech Command task, LUT-NN can even improve accuracy by $1.98\%$ for ResNet18.
\add{Also for the UTKFace dataset, LUT-NN can even reduce MAE by at most 13.2\%.}
Expectedly, the accuracy difference for the more complex tasks, ImageNet and NLP, is a bit larger compared to the tasks above, with $0.9\%$\textasciitilde $2.42\%$ difference for ImageNet, and an averaged $1.9\%$ difference for NLP tasks.

\begin{figure}[htb!]
\centering
\vspace{-0.15in}
\includegraphics[width=0.58\linewidth]{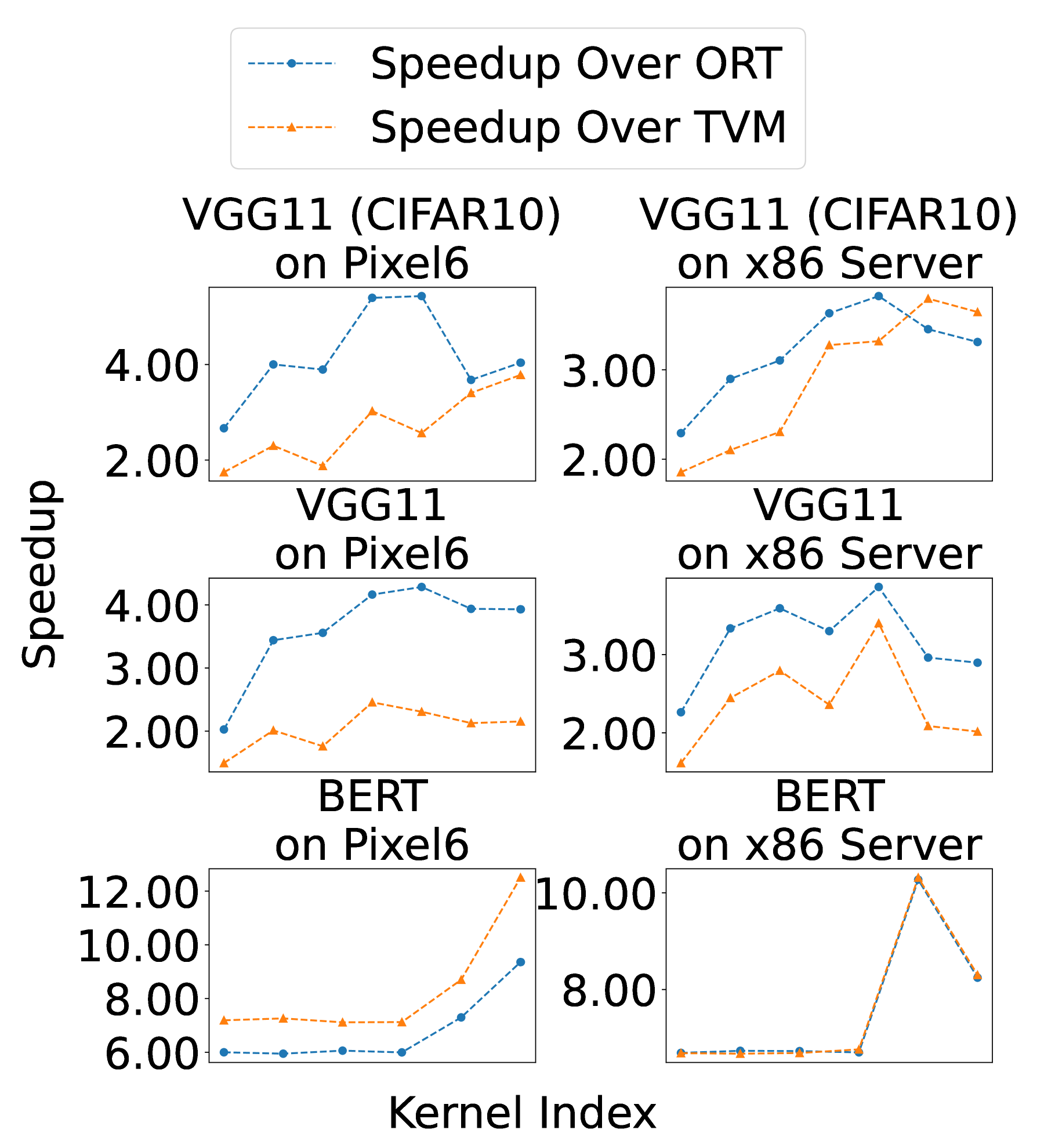}
\vspace{-0.12in}
\caption{
\label{fig:layerwise}
\sysname operator speedup over ORT and TVM on Pixel 6 and the x86 server CPU.
}
\vspace{-0.22in}
\end{figure}

\noindent\textbf{Operator speedup} Fig.~\ref{fig:layerwise} shows the operator speedup selected from three models. The operator speedup for VGG11 ranges from $4.28\times$ to $5.43\times$ on ARM CPU, and from $3.82\times$ to $3.86\times$ on x86 CPU. The best speedup on BERT kernels are $12.5 \times$ (ARM CPU) and $10.3\times$ (x86 CPU). 

BERT operators achieve higher speedups due to larger $M=768$ or $3072$, and longer sub-vector $V=32$. According to Table \ref{tab:syscost}, the FLOPs of our proposed \sysname is $ND (K + M / V)$, and the FLOPs of a normal MM is $N D M$. 
The FLOPs reduction by \sysname can be calculated by $ M / (K + M / V)$. Therefore, the reduction is particularly large when $M \gg K$ and $V \gg 1$. The kernel speedup gradually improves as the layer index increases for CNN models. 
The reason is that the number of output channels increases from $64$ to $128$, $256$, and $512$ as the layer index increases. 
The increased number of output channels $M$ contributes to the speedup in these layers.

\begin{figure*}[htb!]
\centering
\includegraphics[width=0.65\textwidth]{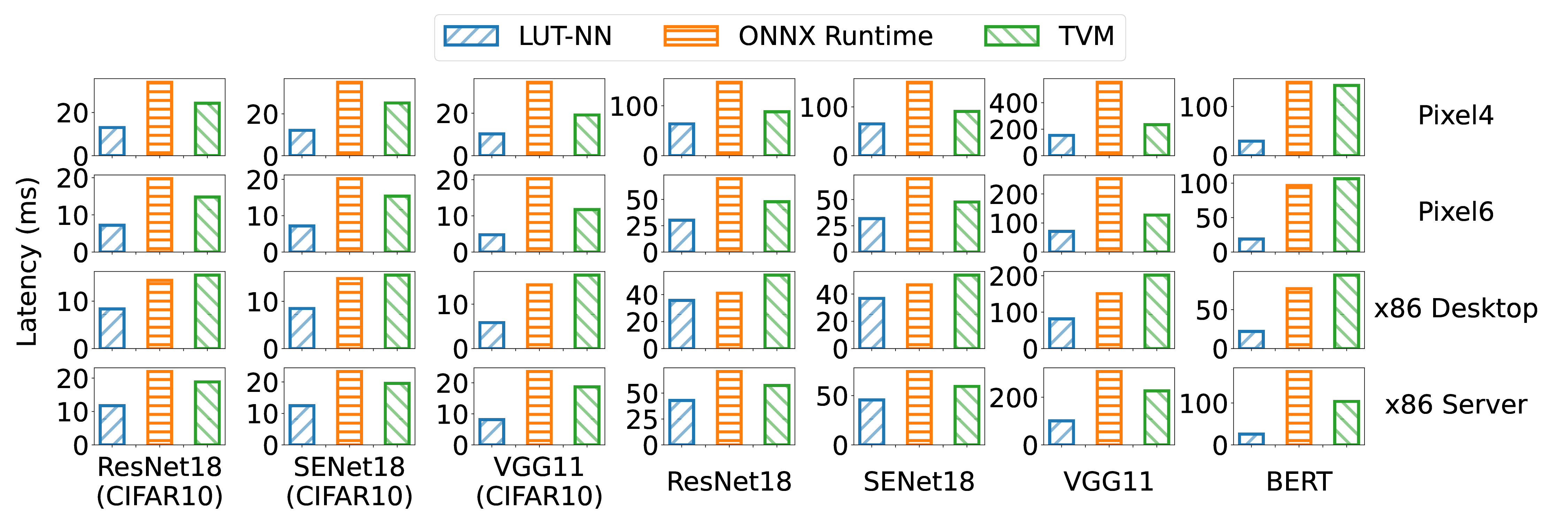}
\vspace{-0.15in}
\caption{
  \label{fig:end_to_end_latency}
  End-to-end latency comparison between \sysname, ORT, and TVM (lower is better).
}
\vspace{-0.2in}
\end{figure*}

\noindent\textbf{Model speedup}
Fig. \ref{fig:end_to_end_latency} shows the end-to-end model latency. 
\sysname achieves $1.30$\textasciitilde$4.23 \times$ speedup on ResNet18, SENet18 and VGG11 compared to TVM and ORT.
The best speedup is $4.23 \times$ in VGG11 (for CIFAR10) on Pixel6 ARM CPU over ONNX Runtime. 
The BERT model has higher throughput, and the speedups are $5.6\times$ and $6.8\times$ on the ARM CPU and the x86 CPU, respectively. 
Compared to CNN models ($M \leq 512$), the BERT model has a larger input tensor ($M = 768, 3072$) to acquire better performance gains. 
It also has a longer sub-vector length ($V = 32$) compared to CNN models ($V \leq 9$).

\begin{figure*}[htb!]
\centering
\includegraphics[width=0.65\linewidth]{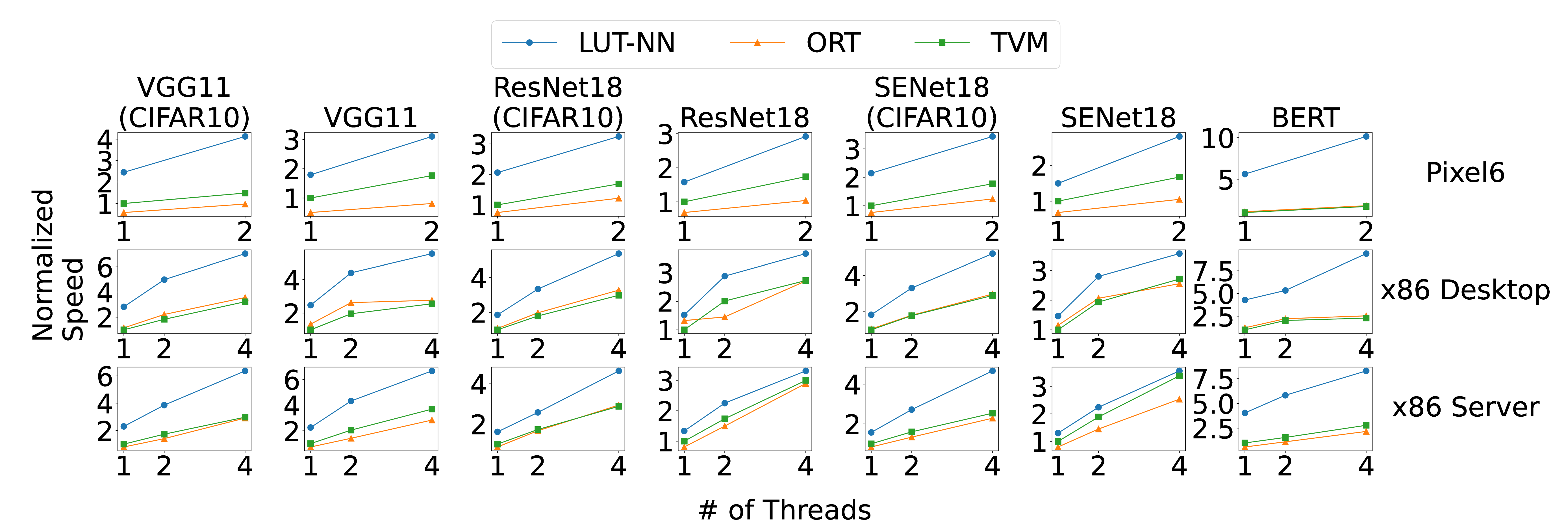}
\vspace{-0.15in}
\caption{
  \add{Normalized multithread speedup (over TVM one thread) of \sysname, ORT, and TVM (higher is better).}
}\label{fig:multi_threading}
\vspace{-0.15in}
\end{figure*}

\noindent\textbf{Multi-thread speedup}
To evaluate the scaling capability of \sysname, Fig.~\ref{fig:multi_threading} shows the speedup with the thread increases, i.e., one to four\add{
on x86 CPUs and one to two on the Pixel 6 ARM CPU. Note that we evaluate the scalability on the big cores (cores with the highest frequency) of ARM CPUs, to avoid the interference of asymmetric big and little cores. Pixel 6 has two big cores. Since Pixel 4 only has one big core, we skip it in this evaluation. 
} The results show that \sysname achieves better scalability compared to the baselines.
It can reach 2.24\textasciitilde 2.49$\times$ speedup as the number of threads increases to 4.
Meanwhile, \sysname achieves a 2.17\textasciitilde 2.82$\times$ speedup over other baselines under the same number of threads.

\begin{figure}[htb!]
\centering
\includegraphics[width=0.6\linewidth]{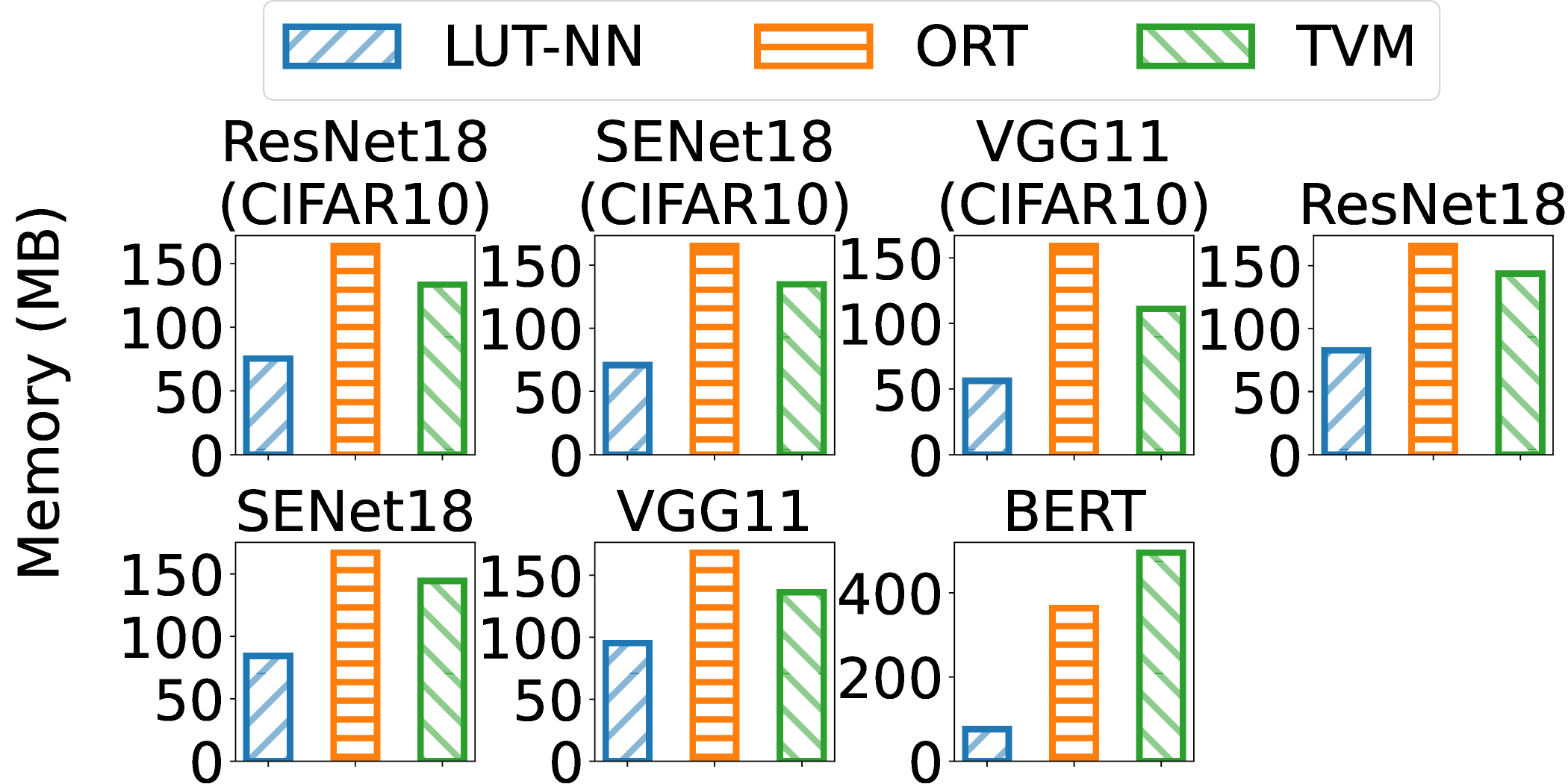}
\vspace{-0.2in}
\caption{
  \label{fig:memory}
  Model memory consumption of \sysname, ORT, and TVM on the x86 server CPU.
}
\vspace{-0.1in}
\end{figure}

\noindent\textbf{Model memory footprint}
Fig.~\ref{fig:memory} compares the memory footprint of \sysname with baselines. 
The results show that \sysname can save $1.43$\textasciitilde$2.83\times$ memory for  CNN models and $4.76$\textasciitilde$6.49\times$ for BERT. For these CNN models, the sub-vector length is $9$ or $4$ for different operators, while the length is set to $16$ for BERT. 
Because longer sub-vectors result in a higher compression ratio on feature maps, BERT achieves lower memory costs than other models.

\noindent\textbf{Power}
We evaluate the power consumption of \sysname on Pixel 4.
The experiment results in Table \ref{tab:end_to_end_power} demonstrate that \sysname reduces power consumption by $15\%$\textasciitilde$41.7\%$  compared to TVM, which reveal that \sysname is an energy-efficient algorithm and system design.


\vspace{-0.15in}
\subsection{Ablation Study}
\label{sec:ablation} 
\vspace{-0.05in}
We evaluate the effectiveness of the learned temperature and hyperparameters in the ablation study.
The hyperparameters include the number of centroids, the length of sub-vector, and the number of replaced linear computation operators. %

\begin{figure}
  \centering
  \includegraphics[width=1\linewidth]{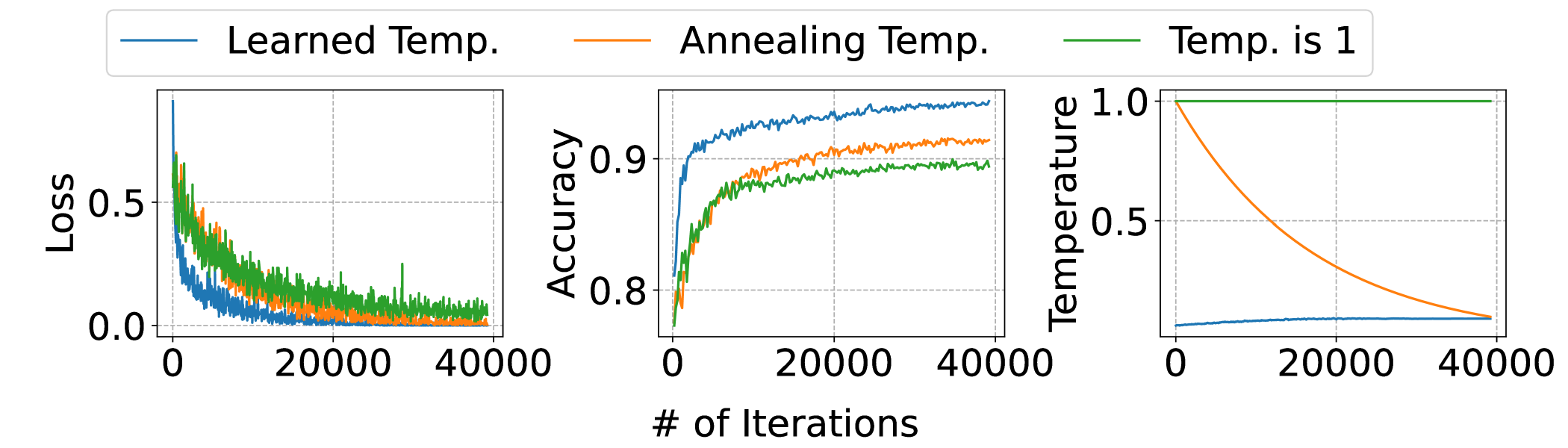}
  \vspace{-0.38in}
  \caption{
  \label{fig:ablation}
  Learning curves of \sysname ResNet18 on CIFAR10 with temperature tuning methods.
  ``Annealing Temperature'' (Orange) refers to manually annealing temperature from 1 to 1e-1.
  Our learned temperature technique reaches 94.4\% accuracy, higher than the annealing temperature (91.55\%) and setting temperature as 1 (89.85\%).
  }
  \vspace{-0.15in}
\end{figure}

\noindent\textbf{Learned temperature} 
In Sec.~\ref{sec:temperature}, we use gradient descent to learn the temperature. 
To evaluate the effectiveness of our method, we compare three temperature tuning strategies together:
learned temperature, statically setting the temperature as $1$, and annealing temperature from $1$ to $0.1$. 
We compare the training curves in Fig.~\ref{fig:ablation}. 
The figure suggests that our proposed learned temperature reaches the highest accuracy ($94.4\%$) and outperforms the statically setting ($89.85\%$) and the annealing temperature ($91.55\%$). 
In addition, the learned temperature has a faster convergence speed.

\noindent\textbf{Impact of centroid number and sub-vector length} 
The centroid number and sub-vector length do affect not only the inference throughput but also the model accuracy. 
We present the ablation study for the impact of these two hyperparameters. 
We follow the same experiment setting as Sec.~\ref{sec:exper_method} on ResNet18.

\begin{figure}[!hbt]
    \vspace{-0.1in}
    \centering
	\includegraphics[width=0.25\textwidth]{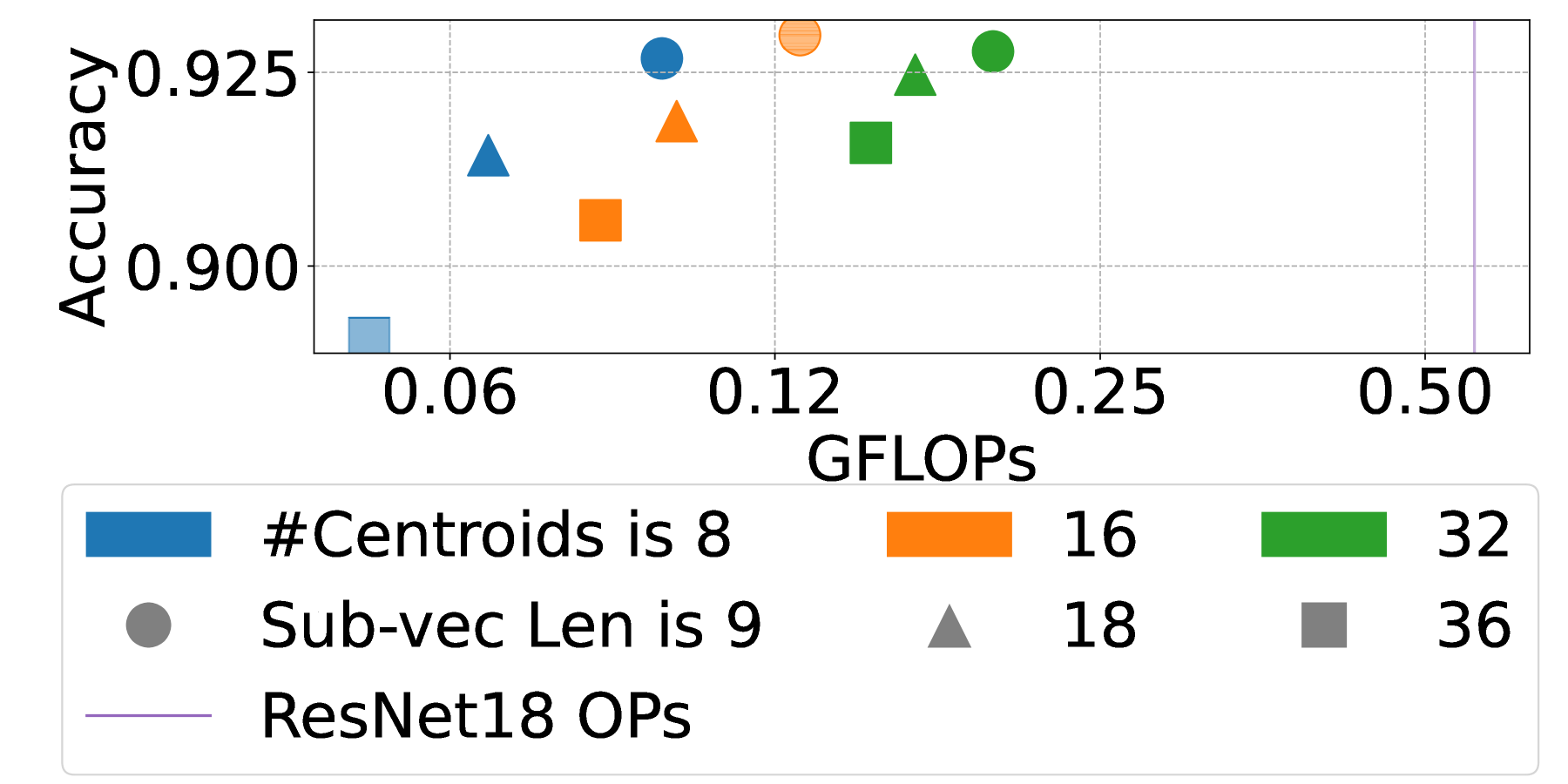}
 \vspace{-0.15in}
 \caption{
 \label{fig:resnet18_grid_search}  
 The scaling of centroid number and sub-vector length on ResNet18. %
 x-axis is logarithmic.
 }
 \vspace{-0.3in}
\end{figure}

Figs.~\ref{fig:resnet18_grid_search} collect accuracy and FLOPs for the variant of centroid numbers and sub-vector lengths in ResNet18.  
For this model, the sub-vector length significantly affects the model accuracy and worsens as the sub-vector length increases. 
Each codebook has to handle higher dimensions as the sub-vector length grows, which cannot be accurately classified and harms the model accuracy. 

The centroid number also affects accuracy and performance. 
As the centroid number increases, each codebook can classify sub-vector in more fine-grained granularity, and it improves the accuracy of ResNet18.
To balance the model accuracy with performance, we set $K = 16$ and $V = 9$ for better accuracy and fewer GFLOPs than the vanilla ResNet18.

\add{
The optimal hyperparameter setting for DNNs is a long-standing research topic. Some neural network search techniques, such as weight sharing~\cite{cai2019once} and search space search~\cite{radosavovic2020designing}, could apply to LUT-NN to reduce the searching cost.
}

\begin{figure}[!hbt]
    \vspace{-0.1in}
    \centering
    \includegraphics[width=0.5\linewidth]{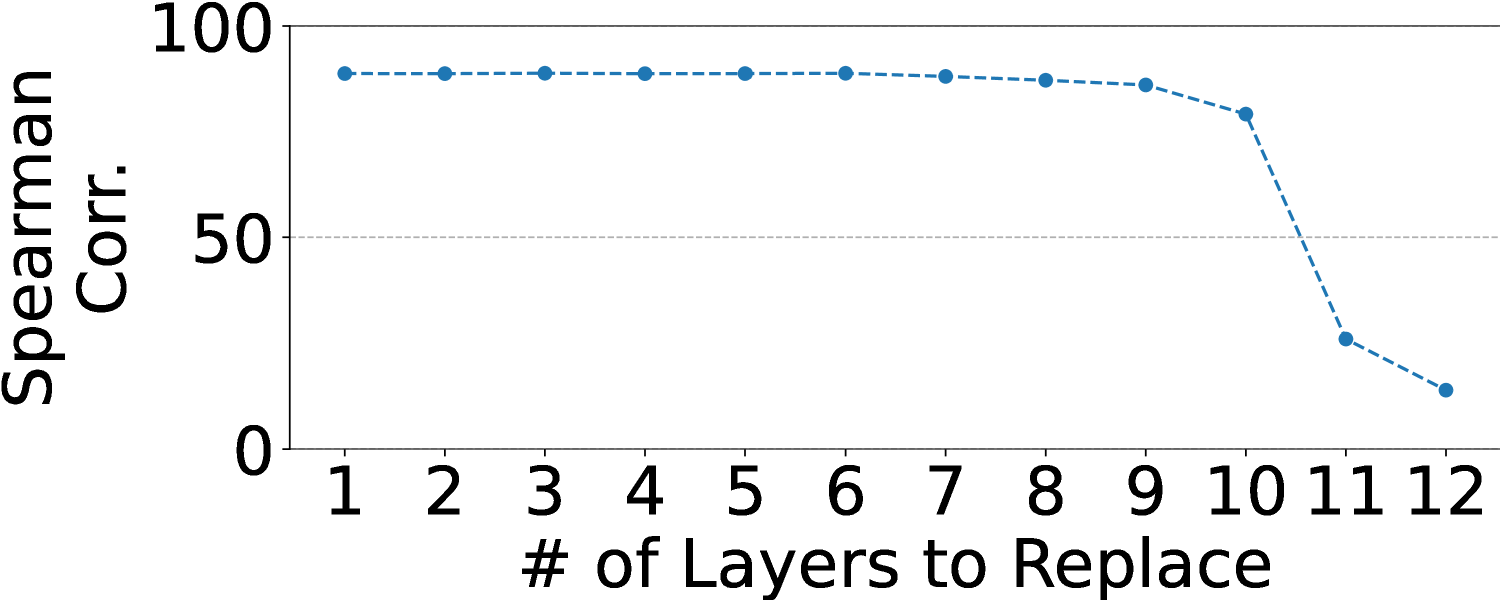}
    \vspace{-0.1in}
    \caption{
    \label{fig:bert_num_layers_to_replace}
    The accuracy of \sysname based BERT with respect to the number of layers to replace. %
    }
    \vspace{-0.25in}
\end{figure}

\noindent\textbf{The number of replaced operators} 
As stated, to replace the front layers of BERT by table lookup can greatly impact accuracy, and we replace the last six layers in our default settings. Fig.~\ref{fig:bert_num_layers_to_replace} shows the accuracy impact by replacing more layers.   
We gradually replace the operators of more layers, from the last to the front layer, for BERT on the Semantic Textual Similarity Benchmark (STS-B) task. 
We can see the accuracy drops obviously for the front three layers. 

\noindent\textbf{Speedup breakdown} 
In Sec. \ref{sec:inference_design}, we propose four inference optimization techniques, which include: \ding{182} memory optimization on centroid distance computations, \ding{183} intra-codebook parallelism, \ding{184} shuffle instructions table lookup, and \ding{185} mix-precision accumulation in Fig. \ref{fig:operator_design}. 
To better understand performance gains, we evaluate the execution of a convolution operator ($C_{in}=C_{out}=64, k=3, s=1, H=W=56$, the second layer of ResNet18) on Pixel 6 when different techniques are enabled. 
The results show that \ding{184} ($44.6\%$) saves the most execution time, followed by methods \ding{182} ($18.5\%$) and \ding{183} ($16.4\%$). Finally, method \ding{185} delivers a minor improvement ($4.1\%$). 
The results show that our inference design and optimization can take advantage of ARM/x86 ISA's special instructions, which outperform conventional operators.

\add{
\noindent\textbf{Scalar quantization level for lookup table}
As explained in Sec.~\ref{scalar-quantized lut}, \sysname leverages scalar quantization to reduce the table size. In this subsection, we evaluate the accuracy impact of different quantization levels by using ResNet18 for the CIFAR10 dataset as an example.    
The accuracy of LUT-NN is 94.44\% with FP32 lookup table, 94.40\% with INT8 lookup table, and 94.27\% with INT4 lookup table. The results show that our quantization-aware training can achieve similar accuracy by using lower bits compared to FP32. We select INT8 in our current implementation as its the tradeoff of accuracy and speedup. INT4, on the other hand, is not supported by the hardware SIMD units directly. The exploration of lower-bit lookup table could be a potential future work.    
}

\begin{table}[!htb]
\centering
\vspace{-0.15in}
\begin{tabular}{|c|c|}
\hline
Model & LUT-NN v.s. TVM Avg. power (W) \\ \hline
BERT & 2.6/3.7 \\ \hline
ResNet18 & 2.6/3.0 \\ \hline
ResNet18 (CIFAR) & 2.6/3.3 \\ \hline
SENET18 & 2.6/2.9 \\ \hline
SENET18 (CIFAR) & 2.8/3.2 \\ \hline
VGG11 & 2.3/2.9 \\ \hline
VGG11 (CIFAR) & 2.7/3.3 \\ \hline
\end{tabular}
\caption{Power (deducted SoC idle power) on Pixel 4.\label{tab:end_to_end_power}}
\vspace{-0.4in}
\end{table}

\vspace{-0.05in}
\section{Related Work}
\vspace{-0.05in}

\noindent\textbf{Approximated matrix multiplication} Traditional approximated matrix multiplication techniques focus on minimizing the difference between the ground truth and the approximated output. 
For example, Magen et al. \cite{magen2011low} use random projection to map two matrices to a lower-dimensional subspace, and then perform accurate multiplication on the mapped matrices. 
\textsc{Maddness}~\cite{MMWM} uses product quantization to learn the most typical features (also known as centroids) in the input distribution. \textsc{Maddness} then stores the products of centroids and the weight matrix in advance in a lookup table, so that the product of any input matrix and weight matrix can be approximated by the result stored in the table.
To conclude, traditional methods will suffer from severe accuracy loss when used in neural network inference if the approximation is too harsh.
Our method can avoid this defect by integrating itself into the end-to-end training process.

Also, previously, some attempts have been made to approximate the weights of NN with low-rank tensor expansion \cite{jaderberg2014speeding, denil2013predicting, denton2014exploiting}.
Specifically, LCNN~\cite{bagherinezhad2017lcnn} is also a kind of low-rank approximation.
With LCNN, a dictionary is trained and the NN weights can be represented by linear combinations of the dictionary items.
Afterward, LCNN transforms a vanilla convolution into lookup operations in the dictionary.
Our work is different from LCNN in that our work learns centroids for each layer's input feature map but not the weight.

\noindent\textbf{Product quantization for DNN}
Product Quantization \cite{matsui2018survey} is a popular and successful method for large-scale approximated nearest neighbor search \cite{jegou2010product}.
In recent years, researchers have integrated product quantization into neural networks for various purposes. 
For example, \cite{yu2018product, yue2016deep, jain2017subic, klein2017defense} incorporated product quantization as a layer in a convolutional neural network to obtain a compact and discriminative image representation. 
\cite{chen2020differentiable} utilized Product Quantization to compress the embedding layer in NLP models. 
\cite{stock2020and, gong2014compressing, chen2020towards} compressed the weight matrix for neural networks with Product Quantization.
However, our approach differs from the above methods that our purpose is to achieve end-to-end neural network inference acceleration.

\noindent\textbf{Scalar quantization}
Scalar quantization methods aim to reduce scalar bits in neural networks, e.g., from float representation to INT8 representation.
They can both compress the neural network bits and accelerate the inference.
Jacob et al. \cite{jacob2018quantization} proposed an INT8 quantization method with both a training scheme and an efficient kernel implementation. 
Other researchers~\cite{courbariaux2015binaryconnect, zhou2016dorefa, alemdar2017ternary, ANT, banner2019post, mckinstry2018discovering} further proved that 1/2/4-bit quantization is sufficient for image classification.
More recently, Wang et al. \cite{wang2022learnable} proposes a new approach to quantization using trainable lookup tables which can fit the distributions in different layers and have small additional computational costs.
Although we leverage scalar quantization to compress the size of the lookup table (Sec.~\ref{scalar-quantized lut}), the main idea of our work is different from these quantization methods.
Our work focuses on learning the most typical features of each NN layer and computing the results of these features in advance for future input tensors, in order to both compress NN weights and accelerate NN inference.

\add{
\noindent\textbf{Inference result reuse}
There are a range of works that exploit the duplicated inputs, to reuse their inference results and   
skip further model inference. 
Cachier~\cite{drolia2017cachier} reuses results among nearby users. Potluck~\cite{guo2018potluck} reuses results across applications. Euphrate~\cite{zhu2018euphrates} reuses results of temporal pixel motion for continuous vision. FoggyCache~\cite{guo2018foggycache} reuses results across devices. DeepCache reuses results of similar regions in recent frames of consecutive videos. MCDNN~\cite{han2016mcdnn} reuses results across applications. Semantic Memory~\cite{li2021boosting} reuses results from some layers for continuous vision.
These works rely on specific application inputs. 
Our work, by comparison, optimizes the model itself to reduce size and FLOPS.%

\noindent\textbf{Lookup table for DNN}
Different from our work that learns and stores the centroid results in the table, there are works that store other results in the tables. %
LogicNets~\cite{umuroglu2020logicnets} maps artificial neurons to truth tables in FPGA. It also compresses the model by available pruning and scalar quantization techniques. Ramanathan et al.~\cite{RamanathanKSCPO20} store the results of bitline computing in Processing in Cache hardware in the table, to accelerate DNN. These works do not target neural network optimization or training techniques like \sysname.    
}

\vspace{-0.12in}
\section{Future opportunity discussion}
\label{sec:gaps}
\label{sec:hashing}
\vspace{-0.05in}


\noindent\textbf{Hardware implication } Potentially, \sysname can greatly reduce FLOPs, e.g., $16\times$ for BERT. However, current hardware restrains \sysname from showing its full potential. 
The reason is that current hardware is optimized for linear computation, but not table lookup. Compared to direct Multiply-Add support for MM in hardware,  
\sysname has to run \verb|argmin| partially sequentially to return the index of the nearest centroid, and then lookup in the table followed by aggregation. 
Besides, the SIMD width limits the options of the number of centroids. These are obviously inefficient. 

Even on this unfriendly hardware, \sysname has shown the benefits of table lookup on all dimensions, particularly for latency, power, and memory reduction. An accelerator or function unit supporting the first-class parallel table-lookup pipeline could even more significantly reduce the inference cost. \sysname opens new opportunity for hardware design. 

\noindent\textbf{Replace more operator types }
Currently \sysname replaces linear computation operators by table lookup, including MM, convolution, and fully connected. The scaled dot-product attention ($<2\%$ of total latency) in attention layers have no weights, and cannot precompute the results with centroids to save in tables, so we do not replace it by lookup table in this work. A possible way to replace it by lookup table is to precompute and save the production results of centroids. In addition, how to replace the non-linear operators, such as the activation functions, by \sysname can be future work. 

\noindent\textbf{Learning for hashing } 
As introduced in Sec.~\ref{sec:pq}, hashing can be used to avoid the Euclidean distance calculation of $k$-means encoding, but at the cost of higher quantization error (refer to Fig.~\ref{fig:ammfordnn-hash}). 
We evaluate the potential of hashing for DNN inference. Since hashing is not differentiable, \sysname uses hashing after the centroids are learned. Our current results show that to achieve similar model accuracy as Euclidean distance for encoding, we have to use a 12-level decision tree for hashing. 
Compared to distance calculation, it can further reduce FLOPs by 30\% to $3\times$. It could be possible to also integrate hashing into the backpropagation to learn the hashing functions. This could reduce the depth of the tree and further reduce the encoding cost.  

On current commodity hardware without direct hashing support, hashing costs even higher than Euclidean distance calculation, since the tree traversing is sequential without SIMD support. Therefore, we still use Euclidean distance for encoding in \sysname now. Future hardware can also integrate hashing units for further speedup.  

\add{
\noindent\textbf{LUT-NN on GPUs and FPGAs}
NVIDIA GPUs provide the efficient shuffle instruction, i.e., “prmt”, achieving Tera-OPs throughput. Similar as the LUT-NN implementation on the CPUs, which heavily depends on the shuffle instructions, this GPU shuffle instruction can be leveraged for parallelly looking up the approximated computation results in the tables.
FPGAs contain a vast array of lookup tables implemented through reconfigurable memory, which LUT-NN can utilize for the storage and retrieval of codebooks. Additionally, FPGAs are equipped with Block RAM, a primary memory resource that enables efficient data storage. It can be employed for LUT-NN’s lookup tables.
}

\vspace{-0.15in}
\section{Conclusion}
\vspace{-0.05in}
This paper takes the first step towards DNN inference by table lookup. A new paradigm potentially brings significant benefits to the DNN inference ecosystem, simplifying the inference software and hardware design and decoupling with the DNN algorithm updates. By the centroid learning technique for DNN, \sysname achieves comparable accuracy for complex tasks with much less resource cost. 
However, \sysname still has space to be improved, such as the learning technique to improve accuracy and hardware support for table lookup, which calls for future efforts from the community. 

\vspace{-0.15in}
\section{Acknowledgement}
\vspace{-0.05in}
The authors thank the anonymous shepherd and reviewers for their valuable comments. The work of Yunxin Liu was supported by the National Key R\&D Program of China (No. 2022YFF0604501) and Xiaomi Foundation. The work of Deng Cai was supported by the National Nature Science Foundation of China (Grant Nos: 62273302, 62036009, 61936006).

\bibliographystyle{ACM-Reference-Format}
\bibliography{references}


\begin{thebibliography}{63}


\ifx \showCODEN    \undefined \def \showCODEN     #1{\unskip}     \fi
\ifx \showDOI      \undefined \def \showDOI       #1{#1}\fi
\ifx \showISBNx    \undefined \def \showISBNx     #1{\unskip}     \fi
\ifx \showISBNxiii \undefined \def \showISBNxiii  #1{\unskip}     \fi
\ifx \showISSN     \undefined \def \showISSN      #1{\unskip}     \fi
\ifx \showLCCN     \undefined \def \showLCCN      #1{\unskip}     \fi
\ifx \shownote     \undefined \def \shownote      #1{#1}          \fi
\ifx \showarticletitle \undefined \def \showarticletitle #1{#1}   \fi
\ifx \showURL      \undefined \def \showURL       {\relax}        \fi
\providecommand\bibfield[2]{#2}
\providecommand\bibinfo[2]{#2}
\providecommand\natexlab[1]{#1}
\providecommand\showeprint[2][]{arXiv:#2}

\bibitem[Alemdar et~al\mbox{.}(2017)]%
        {alemdar2017ternary}
\bibfield{author}{\bibinfo{person}{Hande Alemdar}, \bibinfo{person}{Vincent Leroy}, \bibinfo{person}{Adrien Prost-Boucle}, {and} \bibinfo{person}{Fr{\'e}d{\'e}ric P{\'e}trot}.} \bibinfo{year}{2017}\natexlab{}.
\newblock \showarticletitle{Ternary neural networks for resource-efficient AI applications}. In \bibinfo{booktitle}{\emph{2017 international joint conference on neural networks (IJCNN)}}. IEEE, \bibinfo{pages}{2547--2554}.
\newblock


\bibitem[Bagherinezhad et~al\mbox{.}(2017)]%
        {bagherinezhad2017lcnn}
\bibfield{author}{\bibinfo{person}{Hessam Bagherinezhad}, \bibinfo{person}{Mohammad Rastegari}, {and} \bibinfo{person}{Ali Farhadi}.} \bibinfo{year}{2017}\natexlab{}.
\newblock \showarticletitle{Lcnn: Lookup-based convolutional neural network}. In \bibinfo{booktitle}{\emph{Proceedings of the IEEE conference on computer vision and pattern recognition}}. \bibinfo{pages}{7120--7129}.
\newblock


\bibitem[Banner et~al\mbox{.}(2019)]%
        {banner2019post}
\bibfield{author}{\bibinfo{person}{Ron Banner}, \bibinfo{person}{Yury Nahshan}, {and} \bibinfo{person}{Daniel Soudry}.} \bibinfo{year}{2019}\natexlab{}.
\newblock \showarticletitle{Post training 4-bit quantization of convolutional networks for rapid-deployment}.
\newblock \bibinfo{journal}{\emph{Advances in Neural Information Processing Systems}}  \bibinfo{volume}{32} (\bibinfo{year}{2019}).
\newblock


\bibitem[Blalock et~al\mbox{.}(2020)]%
        {blalock2020state}
\bibfield{author}{\bibinfo{person}{Davis Blalock}, \bibinfo{person}{Jose~Javier Gonzalez~Ortiz}, \bibinfo{person}{Jonathan Frankle}, {and} \bibinfo{person}{John Guttag}.} \bibinfo{year}{2020}\natexlab{}.
\newblock \showarticletitle{What is the state of neural network pruning?}
\newblock \bibinfo{journal}{\emph{Proceedings of machine learning and systems}}  \bibinfo{volume}{2} (\bibinfo{year}{2020}), \bibinfo{pages}{129--146}.
\newblock


\bibitem[Blalock and Guttag(2021)]%
        {MMWM}
\bibfield{author}{\bibinfo{person}{Davis Blalock} {and} \bibinfo{person}{John Guttag}.} \bibinfo{year}{2021}\natexlab{}.
\newblock \showarticletitle{Multiplying Matrices Without Multiplying}. In \bibinfo{booktitle}{\emph{Proceedings of the 38th International Conference on Machine Learning}} \emph{(\bibinfo{series}{Proceedings of Machine Learning Research}, Vol.~\bibinfo{volume}{139})}, \bibfield{editor}{\bibinfo{person}{Marina Meila} {and} \bibinfo{person}{Tong Zhang}} (Eds.). \bibinfo{publisher}{PMLR}, \bibinfo{pages}{992--1004}.
\newblock
\urldef\tempurl%
\url{https://proceedings.mlr.press/v139/blalock21a.html}
\showURL{%
\tempurl}


\bibitem[Blalock and Guttag(2017)]%
        {blalock2017bolt}
\bibfield{author}{\bibinfo{person}{Davis~W Blalock} {and} \bibinfo{person}{John~V Guttag}.} \bibinfo{year}{2017}\natexlab{}.
\newblock \showarticletitle{Bolt: Accelerated data mining with fast vector compression}. In \bibinfo{booktitle}{\emph{Proceedings of the 23rd ACM SIGKDD International Conference on Knowledge Discovery and Data Mining}}. \bibinfo{pages}{727--735}.
\newblock


\bibitem[Cai et~al\mbox{.}(2019)]%
        {cai2019once}
\bibfield{author}{\bibinfo{person}{Han Cai}, \bibinfo{person}{Chuang Gan}, \bibinfo{person}{Tianzhe Wang}, \bibinfo{person}{Zhekai Zhang}, {and} \bibinfo{person}{Song Han}.} \bibinfo{year}{2019}\natexlab{}.
\newblock \showarticletitle{Once-for-all: Train one network and specialize it for efficient deployment}.
\newblock \bibinfo{journal}{\emph{arXiv preprint arXiv:1908.09791}} (\bibinfo{year}{2019}).
\newblock


\bibitem[Chen et~al\mbox{.}(2020a)]%
        {chen2020differentiable}
\bibfield{author}{\bibinfo{person}{Ting Chen}, \bibinfo{person}{Lala Li}, {and} \bibinfo{person}{Yizhou Sun}.} \bibinfo{year}{2020}\natexlab{a}.
\newblock \showarticletitle{Differentiable product quantization for end-to-end embedding compression}. In \bibinfo{booktitle}{\emph{International Conference on Machine Learning}}. PMLR, \bibinfo{pages}{1617--1626}.
\newblock


\bibitem[Chen et~al\mbox{.}(2018)]%
        {chen2018tvm}
\bibfield{author}{\bibinfo{person}{Tianqi Chen}, \bibinfo{person}{Thierry Moreau}, \bibinfo{person}{Ziheng Jiang}, \bibinfo{person}{Lianmin Zheng}, \bibinfo{person}{Eddie Yan}, \bibinfo{person}{Haichen Shen}, \bibinfo{person}{Meghan Cowan}, \bibinfo{person}{Leyuan Wang}, \bibinfo{person}{Yuwei Hu}, \bibinfo{person}{Luis Ceze}, {et~al\mbox{.}}} \bibinfo{year}{2018}\natexlab{}.
\newblock \showarticletitle{TVM: An automated end-to-end optimizing compiler for deep learning}. In \bibinfo{booktitle}{\emph{13th USENIX Symposium on Operating Systems Design and Implementation (OSDI 18)}}. \bibinfo{pages}{578--594}.
\newblock


\bibitem[Chen et~al\mbox{.}(2020b)]%
        {chen2020towards}
\bibfield{author}{\bibinfo{person}{Weihan Chen}, \bibinfo{person}{Peisong Wang}, {and} \bibinfo{person}{Jian Cheng}.} \bibinfo{year}{2020}\natexlab{b}.
\newblock \showarticletitle{Towards Convolutional Neural Networks Compression via Global\&Progressive Product Quantization.}. In \bibinfo{booktitle}{\emph{BMVC}}.
\newblock


\bibitem[Courbariaux et~al\mbox{.}(2015)]%
        {courbariaux2015binaryconnect}
\bibfield{author}{\bibinfo{person}{Matthieu Courbariaux}, \bibinfo{person}{Yoshua Bengio}, {and} \bibinfo{person}{Jean-Pierre David}.} \bibinfo{year}{2015}\natexlab{}.
\newblock \showarticletitle{Binaryconnect: Training deep neural networks with binary weights during propagations}.
\newblock \bibinfo{journal}{\emph{Advances in neural information processing systems}}  \bibinfo{volume}{28} (\bibinfo{year}{2015}).
\newblock


\bibitem[Deng et~al\mbox{.}(2009)]%
        {imagenet}
\bibfield{author}{\bibinfo{person}{Jia Deng}, \bibinfo{person}{Wei Dong}, \bibinfo{person}{Richard Socher}, \bibinfo{person}{Li{-}Jia Li}, \bibinfo{person}{Kai Li}, {and} \bibinfo{person}{Li Fei{-}Fei}.} \bibinfo{year}{2009}\natexlab{}.
\newblock \showarticletitle{ImageNet: {A} large-scale hierarchical image database}. In \bibinfo{booktitle}{\emph{2009 {IEEE} Computer Society Conference on Computer Vision and Pattern Recognition {(CVPR} 2009), 20-25 June 2009, Miami, Florida, {USA}}}. \bibinfo{publisher}{{IEEE} Computer Society}, \bibinfo{pages}{248--255}.
\newblock
\urldef\tempurl%
\url{https://doi.org/10.1109/CVPR.2009.5206848}
\showDOI{\tempurl}


\bibitem[Denil et~al\mbox{.}(2013)]%
        {denil2013predicting}
\bibfield{author}{\bibinfo{person}{Misha Denil}, \bibinfo{person}{Babak Shakibi}, \bibinfo{person}{Laurent Dinh}, \bibinfo{person}{Marc'Aurelio Ranzato}, {and} \bibinfo{person}{Nando De~Freitas}.} \bibinfo{year}{2013}\natexlab{}.
\newblock \showarticletitle{Predicting parameters in deep learning}.
\newblock \bibinfo{journal}{\emph{Advances in neural information processing systems}}  \bibinfo{volume}{26} (\bibinfo{year}{2013}).
\newblock


\bibitem[Denton et~al\mbox{.}(2014)]%
        {denton2014exploiting}
\bibfield{author}{\bibinfo{person}{Emily~L Denton}, \bibinfo{person}{Wojciech Zaremba}, \bibinfo{person}{Joan Bruna}, \bibinfo{person}{Yann LeCun}, {and} \bibinfo{person}{Rob Fergus}.} \bibinfo{year}{2014}\natexlab{}.
\newblock \showarticletitle{Exploiting linear structure within convolutional networks for efficient evaluation}.
\newblock \bibinfo{journal}{\emph{Advances in neural information processing systems}}  \bibinfo{volume}{27} (\bibinfo{year}{2014}).
\newblock


\bibitem[Devlin et~al\mbox{.}(2018)]%
        {devlin2018bert}
\bibfield{author}{\bibinfo{person}{Jacob Devlin}, \bibinfo{person}{Ming-Wei Chang}, \bibinfo{person}{Kenton Lee}, {and} \bibinfo{person}{Kristina Toutanova}.} \bibinfo{year}{2018}\natexlab{}.
\newblock \showarticletitle{Bert: Pre-training of deep bidirectional transformers for language understanding}.
\newblock \bibinfo{journal}{\emph{arXiv preprint arXiv:1810.04805}} (\bibinfo{year}{2018}).
\newblock


\bibitem[Drolia et~al\mbox{.}(2017)]%
        {drolia2017cachier}
\bibfield{author}{\bibinfo{person}{Utsav Drolia}, \bibinfo{person}{Katherine Guo}, \bibinfo{person}{Jiaqi Tan}, \bibinfo{person}{Rajeev Gandhi}, {and} \bibinfo{person}{Priya Narasimhan}.} \bibinfo{year}{2017}\natexlab{}.
\newblock \showarticletitle{Cachier: Edge-caching for recognition applications}. In \bibinfo{booktitle}{\emph{2017 IEEE 37th international conference on distributed computing systems (ICDCS)}}. IEEE, \bibinfo{pages}{276--286}.
\newblock


\bibitem[Gong et~al\mbox{.}(2014)]%
        {gong2014compressing}
\bibfield{author}{\bibinfo{person}{Yunchao Gong}, \bibinfo{person}{Liu Liu}, \bibinfo{person}{Ming Yang}, {and} \bibinfo{person}{Lubomir Bourdev}.} \bibinfo{year}{2014}\natexlab{}.
\newblock \showarticletitle{Compressing deep convolutional networks using vector quantization}.
\newblock \bibinfo{journal}{\emph{arXiv preprint arXiv:1412.6115}} (\bibinfo{year}{2014}).
\newblock


\bibitem[Google(2019)]%
        {tensorflow}
\bibfield{author}{\bibinfo{person}{Google}.} \bibinfo{year}{2019}\natexlab{}.
\newblock \bibinfo{booktitle}{\emph{TensorFlow: An end-to-end open source machine learning platform}}.
\newblock
\urldef\tempurl%
\url{https://www.tensorflow.org/}
\showURL{%
\tempurl}


\bibitem[Gou et~al\mbox{.}(2021)]%
        {gou2021knowledge}
\bibfield{author}{\bibinfo{person}{Jianping Gou}, \bibinfo{person}{Baosheng Yu}, \bibinfo{person}{Stephen~J Maybank}, {and} \bibinfo{person}{Dacheng Tao}.} \bibinfo{year}{2021}\natexlab{}.
\newblock \showarticletitle{Knowledge distillation: A survey}.
\newblock \bibinfo{journal}{\emph{International Journal of Computer Vision}}  \bibinfo{volume}{129} (\bibinfo{year}{2021}), \bibinfo{pages}{1789--1819}.
\newblock


\bibitem[Gray and Neuhoff(1998)]%
        {gray1998quantization}
\bibfield{author}{\bibinfo{person}{R.M. Gray} {and} \bibinfo{person}{D.L. Neuhoff}.} \bibinfo{year}{1998}\natexlab{}.
\newblock \showarticletitle{Quantization}.
\newblock \bibinfo{journal}{\emph{IEEE Transactions on Information Theory}} \bibinfo{volume}{44}, \bibinfo{number}{6} (\bibinfo{year}{1998}), \bibinfo{pages}{2325--2383}.
\newblock
\urldef\tempurl%
\url{https://doi.org/10.1109/18.720541}
\showDOI{\tempurl}


\bibitem[Guo et~al\mbox{.}(2022)]%
        {ANT}
\bibfield{author}{\bibinfo{person}{Cong Guo}, \bibinfo{person}{Chen Zhang}, \bibinfo{person}{Jingwen Leng}, \bibinfo{person}{Zihan Liu}, \bibinfo{person}{Fan Yang}, \bibinfo{person}{Yunxin Liu}, \bibinfo{person}{Minyi Guo}, {and} \bibinfo{person}{Yuhao Zhu}.} \bibinfo{year}{2022}\natexlab{}.
\newblock \showarticletitle{{ANT:} Exploiting Adaptive Numerical Data Type for Low-bit Deep Neural Network Quantization}. In \bibinfo{booktitle}{\emph{55th {IEEE/ACM} International Symposium on Microarchitecture, {MICRO} 2022, Chicago, IL, USA, October 1-5, 2022}}. \bibinfo{publisher}{{IEEE}}, \bibinfo{pages}{1414--1433}.
\newblock
\urldef\tempurl%
\url{https://doi.org/10.1109/MICRO56248.2022.00095}
\showDOI{\tempurl}


\bibitem[Guo et~al\mbox{.}(2018)]%
        {guo2018foggycache}
\bibfield{author}{\bibinfo{person}{Peizhen Guo}, \bibinfo{person}{Bo Hu}, \bibinfo{person}{Rui Li}, {and} \bibinfo{person}{Wenjun Hu}.} \bibinfo{year}{2018}\natexlab{}.
\newblock \showarticletitle{Foggycache: Cross-device approximate computation reuse}. In \bibinfo{booktitle}{\emph{Proceedings of the 24th annual international conference on mobile computing and networking}}. \bibinfo{pages}{19--34}.
\newblock


\bibitem[Guo and Hu(2018)]%
        {guo2018potluck}
\bibfield{author}{\bibinfo{person}{Peizhen Guo} {and} \bibinfo{person}{Wenjun Hu}.} \bibinfo{year}{2018}\natexlab{}.
\newblock \showarticletitle{Potluck: Cross-application approximate deduplication for computation-intensive mobile applications}. In \bibinfo{booktitle}{\emph{Proceedings of the Twenty-Third International Conference on Architectural Support for Programming Languages and Operating Systems}}. \bibinfo{pages}{271--284}.
\newblock


\bibitem[Guo et~al\mbox{.}(2015)]%
        {GuoKCS15}
\bibfield{author}{\bibinfo{person}{Ruiqi Guo}, \bibinfo{person}{Sanjiv Kumar}, \bibinfo{person}{Krzysztof Choromanski}, {and} \bibinfo{person}{David Simcha}.} \bibinfo{year}{2015}\natexlab{}.
\newblock \showarticletitle{Quantization based Fast Inner Product Search}.
\newblock \bibinfo{journal}{\emph{CoRR}}  \bibinfo{volume}{abs/1509.01469} (\bibinfo{year}{2015}).
\newblock
\showeprint[arXiv]{1509.01469}
\urldef\tempurl%
\url{http://arxiv.org/abs/1509.01469}
\showURL{%
\tempurl}


\bibitem[Han et~al\mbox{.}(2016)]%
        {han2016mcdnn}
\bibfield{author}{\bibinfo{person}{Seungyeop Han}, \bibinfo{person}{Haichen Shen}, \bibinfo{person}{Matthai Philipose}, \bibinfo{person}{Sharad Agarwal}, \bibinfo{person}{Alec Wolman}, {and} \bibinfo{person}{Arvind Krishnamurthy}.} \bibinfo{year}{2016}\natexlab{}.
\newblock \showarticletitle{Mcdnn: An approximation-based execution framework for deep stream processing under resource constraints}. In \bibinfo{booktitle}{\emph{Proceedings of the 14th Annual International Conference on Mobile Systems, Applications, and Services}}. \bibinfo{pages}{123--136}.
\newblock


\bibitem[He et~al\mbox{.}(2013)]%
        {HeWS13}
\bibfield{author}{\bibinfo{person}{Kaiming He}, \bibinfo{person}{Fang Wen}, {and} \bibinfo{person}{Jian Sun}.} \bibinfo{year}{2013}\natexlab{}.
\newblock \showarticletitle{K-Means Hashing: An Affinity-Preserving Quantization Method for Learning Binary Compact Codes}. In \bibinfo{booktitle}{\emph{2013 {IEEE} Conference on Computer Vision and Pattern Recognition, Portland, OR, USA, June 23-28, 2013}}. \bibinfo{publisher}{{IEEE} Computer Society}, \bibinfo{pages}{2938--2945}.
\newblock
\urldef\tempurl%
\url{https://doi.org/10.1109/CVPR.2013.378}
\showDOI{\tempurl}


\bibitem[He et~al\mbox{.}(2016)]%
        {he2016deep}
\bibfield{author}{\bibinfo{person}{Kaiming He}, \bibinfo{person}{Xiangyu Zhang}, \bibinfo{person}{Shaoqing Ren}, {and} \bibinfo{person}{Jian Sun}.} \bibinfo{year}{2016}\natexlab{}.
\newblock \showarticletitle{Deep residual learning for image recognition}. In \bibinfo{booktitle}{\emph{Proceedings of the IEEE conference on computer vision and pattern recognition}}. \bibinfo{pages}{770--778}.
\newblock


\bibitem[Hinton et~al\mbox{.}(2015)]%
        {hinton2015}
\bibfield{author}{\bibinfo{person}{Geoffrey Hinton}, \bibinfo{person}{Oriol Vinyals}, {and} \bibinfo{person}{Jeff Dean}.} \bibinfo{year}{2015}\natexlab{}.
\newblock \bibinfo{title}{Distilling the Knowledge in a Neural Network}.
\newblock
\newblock
\urldef\tempurl%
\url{https://doi.org/10.48550/ARXIV.1503.02531}
\showDOI{\tempurl}


\bibitem[Jacob et~al\mbox{.}(2018)]%
        {jacob2018quantization}
\bibfield{author}{\bibinfo{person}{Benoit Jacob}, \bibinfo{person}{Skirmantas Kligys}, \bibinfo{person}{Bo Chen}, \bibinfo{person}{Menglong Zhu}, \bibinfo{person}{Matthew Tang}, \bibinfo{person}{Andrew Howard}, \bibinfo{person}{Hartwig Adam}, {and} \bibinfo{person}{Dmitry Kalenichenko}.} \bibinfo{year}{2018}\natexlab{}.
\newblock \showarticletitle{Quantization and training of neural networks for efficient integer-arithmetic-only inference}. In \bibinfo{booktitle}{\emph{Proceedings of the IEEE conference on computer vision and pattern recognition}}. \bibinfo{pages}{2704--2713}.
\newblock


\bibitem[Jaderberg et~al\mbox{.}(2014)]%
        {jaderberg2014speeding}
\bibfield{author}{\bibinfo{person}{Max Jaderberg}, \bibinfo{person}{Andrea Vedaldi}, {and} \bibinfo{person}{Andrew Zisserman}.} \bibinfo{year}{2014}\natexlab{}.
\newblock \showarticletitle{Speeding up convolutional neural networks with low rank expansions}.
\newblock \bibinfo{journal}{\emph{arXiv preprint arXiv:1405.3866}} (\bibinfo{year}{2014}).
\newblock


\bibitem[Jain et~al\mbox{.}(2017)]%
        {jain2017subic}
\bibfield{author}{\bibinfo{person}{Himalaya Jain}, \bibinfo{person}{Joaquin Zepeda}, \bibinfo{person}{Patrick P{\'e}rez}, {and} \bibinfo{person}{R{\'e}mi Gribonval}.} \bibinfo{year}{2017}\natexlab{}.
\newblock \showarticletitle{Subic: A supervised, structured binary code for image search}. In \bibinfo{booktitle}{\emph{Proceedings of the IEEE international conference on computer vision}}. \bibinfo{pages}{833--842}.
\newblock


\bibitem[Jang et~al\mbox{.}(2016)]%
        {jang2016}
\bibfield{author}{\bibinfo{person}{Eric Jang}, \bibinfo{person}{Shixiang Gu}, {and} \bibinfo{person}{Ben Poole}.} \bibinfo{year}{2016}\natexlab{}.
\newblock \bibinfo{title}{Categorical Reparameterization with Gumbel-Softmax}.
\newblock
\newblock
\urldef\tempurl%
\url{https://doi.org/10.48550/ARXIV.1611.01144}
\showDOI{\tempurl}


\bibitem[Jegou et~al\mbox{.}(2010)]%
        {jegou2010product}
\bibfield{author}{\bibinfo{person}{Herve Jegou}, \bibinfo{person}{Matthijs Douze}, {and} \bibinfo{person}{Cordelia Schmid}.} \bibinfo{year}{2010}\natexlab{}.
\newblock \showarticletitle{Product quantization for nearest neighbor search}.
\newblock \bibinfo{journal}{\emph{IEEE transactions on pattern analysis and machine intelligence}} \bibinfo{volume}{33}, \bibinfo{number}{1} (\bibinfo{year}{2010}), \bibinfo{pages}{117--128}.
\newblock


\bibitem[Jiao et~al\mbox{.}(2020)]%
        {JiaoHL20}
\bibfield{author}{\bibinfo{person}{Yang Jiao}, \bibinfo{person}{Liang Han}, {and} \bibinfo{person}{Xin Long}.} \bibinfo{year}{2020}\natexlab{}.
\newblock \showarticletitle{Hanguang 800 {NPU} - The Ultimate {AI} Inference Solution for Data Centers}. In \bibinfo{booktitle}{\emph{{IEEE} Hot Chips 32 Symposium, {HCS} 2020, Palo Alto, CA, USA, August 16-18, 2020}}. \bibinfo{publisher}{{IEEE}}, \bibinfo{pages}{1--29}.
\newblock
\urldef\tempurl%
\url{https://doi.org/10.1109/HCS49909.2020.9220619}
\showDOI{\tempurl}


\bibitem[Jouppi et~al\mbox{.}(2021)]%
        {tpuv4}
\bibfield{author}{\bibinfo{person}{Norman~P. Jouppi}, \bibinfo{person}{Doe~Hyun Yoon}, \bibinfo{person}{Matthew Ashcraft}, \bibinfo{person}{Mark Gottscho}, \bibinfo{person}{Thomas~B. Jablin}, \bibinfo{person}{George Kurian}, \bibinfo{person}{James Laudon}, \bibinfo{person}{Sheng Li}, \bibinfo{person}{Peter Ma}, \bibinfo{person}{Xiaoyu Ma}, \bibinfo{person}{Thomas Norrie}, \bibinfo{person}{Nishant Patil}, \bibinfo{person}{Sushma Prasad}, \bibinfo{person}{Cliff Young}, \bibinfo{person}{Zongwei Zhou}, {and} \bibinfo{person}{David Patterson}.} \bibinfo{year}{2021}\natexlab{}.
\newblock \showarticletitle{Ten Lessons from Three Generations Shaped Google's TPUv4i} \emph{(\bibinfo{series}{ISCA '21})}. \bibinfo{publisher}{IEEE Press}, \bibinfo{pages}{1–14}.
\newblock
\urldef\tempurl%
\url{https://doi.org/10.1109/ISCA52012.2021.00010}
\showDOI{\tempurl}


\bibitem[Klein and Wolf(2017)]%
        {klein2017defense}
\bibfield{author}{\bibinfo{person}{Benjamin Klein} {and} \bibinfo{person}{Lior Wolf}.} \bibinfo{year}{2017}\natexlab{}.
\newblock \showarticletitle{In defense of product quantization}.
\newblock \bibinfo{journal}{\emph{arXiv preprint arXiv:1711.08589}} \bibinfo{volume}{2}, \bibinfo{number}{3} (\bibinfo{year}{2017}), \bibinfo{pages}{4}.
\newblock


\bibitem[Krizhevsky(2009)]%
        {cifar}
\bibfield{author}{\bibinfo{person}{Alex Krizhevsky}.} \bibinfo{year}{2009}\natexlab{}.
\newblock \bibinfo{booktitle}{\emph{Learning multiple layers of features from tiny images}}.
\newblock \bibinfo{type}{{T}echnical {R}eport}.
\newblock


\bibitem[Li et~al\mbox{.}(2021)]%
        {li2021boosting}
\bibfield{author}{\bibinfo{person}{Yun Li}, \bibinfo{person}{Chen Zhang}, \bibinfo{person}{Shihao Han}, \bibinfo{person}{Li~Lyna Zhang}, \bibinfo{person}{Baoqun Yin}, \bibinfo{person}{Yunxin Liu}, {and} \bibinfo{person}{Mengwei Xu}.} \bibinfo{year}{2021}\natexlab{}.
\newblock \showarticletitle{Boosting mobile CNN inference through semantic memory}. In \bibinfo{booktitle}{\emph{Proceedings of the 29th ACM International Conference on Multimedia}}. \bibinfo{pages}{2362--2371}.
\newblock


\bibitem[Liang et~al\mbox{.}(2022)]%
        {liang2022romou}
\bibfield{author}{\bibinfo{person}{Rendong Liang}, \bibinfo{person}{Ting Cao}, \bibinfo{person}{Jicheng Wen}, \bibinfo{person}{Manni Wang}, \bibinfo{person}{Yang Wang}, \bibinfo{person}{Jianhua Zou}, {and} \bibinfo{person}{Yunxin Liu}.} \bibinfo{year}{2022}\natexlab{}.
\newblock \showarticletitle{Romou: Rapidly generate high-performance tensor kernels for mobile gpus}. In \bibinfo{booktitle}{\emph{Proceedings of the 28th Annual International Conference on Mobile Computing And Networking}}. \bibinfo{pages}{487--500}.
\newblock


\bibitem[Lloyd(1982)]%
        {Lloyd82}
\bibfield{author}{\bibinfo{person}{Stuart~P. Lloyd}.} \bibinfo{year}{1982}\natexlab{}.
\newblock \showarticletitle{Least squares quantization in {PCM}}.
\newblock \bibinfo{journal}{\emph{{IEEE} Trans. Inf. Theory}} \bibinfo{volume}{28}, \bibinfo{number}{2} (\bibinfo{year}{1982}), \bibinfo{pages}{129--136}.
\newblock
\urldef\tempurl%
\url{https://doi.org/10.1109/TIT.1982.1056489}
\showDOI{\tempurl}


\bibitem[MACE(2020)]%
        {MACE}
\bibfield{author}{\bibinfo{person}{MACE}.} \bibinfo{year}{2020}\natexlab{}.
\newblock
\newblock
\urldef\tempurl%
\url{https://github.com/XiaoMi/mace.}
\showURL{%
\tempurl}


\bibitem[Magen and Zouzias(2011)]%
        {magen2011low}
\bibfield{author}{\bibinfo{person}{Avner Magen} {and} \bibinfo{person}{Anastasios Zouzias}.} \bibinfo{year}{2011}\natexlab{}.
\newblock \showarticletitle{Low rank matrix-valued Chernoff bounds and approximate matrix multiplication}. In \bibinfo{booktitle}{\emph{Proceedings of the twenty-second annual ACM-SIAM symposium on Discrete Algorithms}}. SIAM, \bibinfo{pages}{1422--1436}.
\newblock


\bibitem[Matsui et~al\mbox{.}(2018)]%
        {matsui2018survey}
\bibfield{author}{\bibinfo{person}{Yusuke Matsui}, \bibinfo{person}{Yusuke Uchida}, \bibinfo{person}{Herv{\'e} J{\'e}gou}, {and} \bibinfo{person}{Shin'ichi Satoh}.} \bibinfo{year}{2018}\natexlab{}.
\newblock \showarticletitle{A survey of product quantization}.
\newblock \bibinfo{journal}{\emph{ITE Transactions on Media Technology and Applications}} \bibinfo{volume}{6}, \bibinfo{number}{1} (\bibinfo{year}{2018}), \bibinfo{pages}{2--10}.
\newblock


\bibitem[McCarter and Dronen(2022)]%
        {McCarter2022}
\bibfield{author}{\bibinfo{person}{Calvin McCarter} {and} \bibinfo{person}{Nicholas Dronen}.} \bibinfo{year}{2022}\natexlab{}.
\newblock \showarticletitle{Look-ups are not (yet) all you need for deep learning inference}.
\newblock \bibinfo{journal}{\emph{arXiv preprint arXiv:2207.05808}} (\bibinfo{year}{2022}).
\newblock


\bibitem[McKinstry et~al\mbox{.}(2018)]%
        {mckinstry2018discovering}
\bibfield{author}{\bibinfo{person}{Jeffrey~L McKinstry}, \bibinfo{person}{Steven~K Esser}, \bibinfo{person}{Rathinakumar Appuswamy}, \bibinfo{person}{Deepika Bablani}, \bibinfo{person}{John~V Arthur}, \bibinfo{person}{Izzet~B Yildiz}, {and} \bibinfo{person}{Dharmendra~S Modha}.} \bibinfo{year}{2018}\natexlab{}.
\newblock \showarticletitle{Discovering low-precision networks close to full-precision networks for efficient embedded inference}.
\newblock \bibinfo{journal}{\emph{arXiv preprint arXiv:1809.04191}} (\bibinfo{year}{2018}).
\newblock


\bibitem[Microsoft(2019)]%
        {onnx}
\bibfield{author}{\bibinfo{person}{Microsoft}.} \bibinfo{year}{2019}\natexlab{}.
\newblock \bibinfo{booktitle}{\emph{ONNX Runtime}}.
\newblock
\urldef\tempurl%
\url{https://github.com/microsoft/onnxruntime}
\showURL{%
\tempurl}


\bibitem[Netzer et~al\mbox{.}(2011)]%
        {SVHN}
\bibfield{author}{\bibinfo{person}{Yuval Netzer}, \bibinfo{person}{Tao Wang}, \bibinfo{person}{Adam Coates}, \bibinfo{person}{Alessandro Bissacco}, \bibinfo{person}{Bo Wu}, {and} \bibinfo{person}{Andrew~Y Ng}.} \bibinfo{year}{2011}\natexlab{}.
\newblock \showarticletitle{Reading digits in natural images with unsupervised feature learning}. In \bibinfo{booktitle}{\emph{NIPS Workshop on Deep Learning and Unsupervised Feature Learning 2011}}.
\newblock


\bibitem[Radosavovic et~al\mbox{.}(2020)]%
        {radosavovic2020designing}
\bibfield{author}{\bibinfo{person}{Ilija Radosavovic}, \bibinfo{person}{Raj~Prateek Kosaraju}, \bibinfo{person}{Ross Girshick}, \bibinfo{person}{Kaiming He}, {and} \bibinfo{person}{Piotr Doll{\'a}r}.} \bibinfo{year}{2020}\natexlab{}.
\newblock \showarticletitle{Designing network design spaces}. In \bibinfo{booktitle}{\emph{Proceedings of the IEEE/CVF conference on computer vision and pattern recognition}}. \bibinfo{pages}{10428--10436}.
\newblock


\bibitem[Ramanathan et~al\mbox{.}(2020)]%
        {RamanathanKSCPO20}
\bibfield{author}{\bibinfo{person}{Akshay~Krishna Ramanathan}, \bibinfo{person}{Gurpreet~S. Kalsi}, \bibinfo{person}{Srivatsa Srinivasa}, \bibinfo{person}{Tarun~Makesh Chandran}, \bibinfo{person}{Kamlesh~R. Pillai}, \bibinfo{person}{Om~Ji Omer}, \bibinfo{person}{Vijaykrishnan Narayanan}, {and} \bibinfo{person}{Sreenivas Subramoney}.} \bibinfo{year}{2020}\natexlab{}.
\newblock \showarticletitle{Look-Up Table based Energy Efficient Processing in Cache Support for Neural Network Acceleration}. In \bibinfo{booktitle}{\emph{53rd Annual {IEEE/ACM} International Symposium on Microarchitecture, {MICRO} 2020, Athens, Greece, October 17-21, 2020}}. \bibinfo{publisher}{{IEEE}}, \bibinfo{pages}{88--101}.
\newblock
\urldef\tempurl%
\url{https://doi.org/10.1109/MICRO50266.2020.00020}
\showDOI{\tempurl}


\bibitem[Stallkamp et~al\mbox{.}(2011)]%
        {GTSRB}
\bibfield{author}{\bibinfo{person}{Johannes Stallkamp}, \bibinfo{person}{Marc Schlipsing}, \bibinfo{person}{Jan Salmen}, {and} \bibinfo{person}{Christian Igel}.} \bibinfo{year}{2011}\natexlab{}.
\newblock \showarticletitle{The German Traffic Sign Recognition Benchmark: A multi-class classification competition}. In \bibinfo{booktitle}{\emph{The 2011 International Joint Conference on Neural Networks}}. \bibinfo{pages}{1453--1460}.
\newblock
\urldef\tempurl%
\url{https://doi.org/10.1109/IJCNN.2011.6033395}
\showDOI{\tempurl}


\bibitem[Stock et~al\mbox{.}(2020)]%
        {stock2020and}
\bibfield{author}{\bibinfo{person}{Pierre Stock}, \bibinfo{person}{Armand Joulin}, \bibinfo{person}{R{\'e}mi Gribonval}, \bibinfo{person}{Benjamin Graham}, {and} \bibinfo{person}{Herv{\'e} J{\'e}gou}.} \bibinfo{year}{2020}\natexlab{}.
\newblock \showarticletitle{And the Bit Goes Down: Revisiting the Quantization of Neural Networks}. In \bibinfo{booktitle}{\emph{ICLR 2020-Eighth International Conference on Learning Representations}}. \bibinfo{pages}{1--11}.
\newblock


\bibitem[Umuroglu et~al\mbox{.}(2020)]%
        {umuroglu2020logicnets}
\bibfield{author}{\bibinfo{person}{Yaman Umuroglu}, \bibinfo{person}{Yash Akhauri}, \bibinfo{person}{Nicholas~James Fraser}, {and} \bibinfo{person}{Michaela Blott}.} \bibinfo{year}{2020}\natexlab{}.
\newblock \showarticletitle{LogicNets: Co-designed neural networks and circuits for extreme-throughput applications}. In \bibinfo{booktitle}{\emph{2020 30th International Conference on Field-Programmable Logic and Applications (FPL)}}. IEEE, \bibinfo{pages}{291--297}.
\newblock


\bibitem[Wang et~al\mbox{.}(2018)]%
        {glue}
\bibfield{author}{\bibinfo{person}{Alex Wang}, \bibinfo{person}{Amanpreet Singh}, \bibinfo{person}{Julian Michael}, \bibinfo{person}{Felix Hill}, \bibinfo{person}{Omer Levy}, {and} \bibinfo{person}{Samuel~R. Bowman}.} \bibinfo{year}{2018}\natexlab{}.
\newblock \showarticletitle{{GLUE:} {A} Multi-Task Benchmark and Analysis Platform for Natural Language Understanding}.
\newblock \bibinfo{journal}{\emph{CoRR}}  \bibinfo{volume}{abs/1804.07461} (\bibinfo{year}{2018}).
\newblock
\showeprint[arXiv]{1804.07461}
\urldef\tempurl%
\url{http://arxiv.org/abs/1804.07461}
\showURL{%
\tempurl}


\bibitem[Wang et~al\mbox{.}(2022)]%
        {wang2022learnable}
\bibfield{author}{\bibinfo{person}{Longguang Wang}, \bibinfo{person}{Xiaoyu Dong}, \bibinfo{person}{Yingqian Wang}, \bibinfo{person}{Li Liu}, \bibinfo{person}{Wei An}, {and} \bibinfo{person}{Yulan Guo}.} \bibinfo{year}{2022}\natexlab{}.
\newblock \showarticletitle{Learnable lookup table for neural network quantization}. In \bibinfo{booktitle}{\emph{Proceedings of the IEEE/CVF conference on computer vision and pattern recognition}}. \bibinfo{pages}{12423--12433}.
\newblock


\bibitem[Wang et~al\mbox{.}(2021)]%
        {wang2021asymo}
\bibfield{author}{\bibinfo{person}{Manni Wang}, \bibinfo{person}{Shaohua Ding}, \bibinfo{person}{Ting Cao}, \bibinfo{person}{Yunxin Liu}, {and} \bibinfo{person}{Fengyuan Xu}.} \bibinfo{year}{2021}\natexlab{}.
\newblock \showarticletitle{Asymo: scalable and efficient deep-learning inference on asymmetric mobile cpus}. In \bibinfo{booktitle}{\emph{Proceedings of the 27th Annual International Conference on Mobile Computing and Networking}}. \bibinfo{pages}{215--228}.
\newblock


\bibitem[Warden(2018)]%
        {speech_command}
\bibfield{author}{\bibinfo{person}{Pete Warden}.} \bibinfo{year}{2018}\natexlab{}.
\newblock \showarticletitle{Speech commands: A dataset for limited-vocabulary speech recognition}.
\newblock \bibinfo{journal}{\emph{arXiv preprint arXiv:1804.03209}} (\bibinfo{year}{2018}).
\newblock


\bibitem[Wechsler et~al\mbox{.}(2019)]%
        {nnpi}
\bibfield{author}{\bibinfo{person}{Ofri Wechsler}, \bibinfo{person}{Michael Behar}, {and} \bibinfo{person}{Bharat Daga}.} \bibinfo{year}{2019}\natexlab{}.
\newblock \showarticletitle{Spring Hill (NNP-I 1000) Intel’s Data Center Inference Chip}. In \bibinfo{booktitle}{\emph{2019 IEEE Hot Chips 31 Symposium (HCS)}}. \bibinfo{pages}{1--12}.
\newblock
\urldef\tempurl%
\url{https://doi.org/10.1109/HOTCHIPS.2019.8875671}
\showDOI{\tempurl}


\bibitem[Yu et~al\mbox{.}(2018)]%
        {yu2018product}
\bibfield{author}{\bibinfo{person}{Tan Yu}, \bibinfo{person}{Junsong Yuan}, \bibinfo{person}{Chen Fang}, {and} \bibinfo{person}{Hailin Jin}.} \bibinfo{year}{2018}\natexlab{}.
\newblock \showarticletitle{Product quantization network for fast image retrieval}. In \bibinfo{booktitle}{\emph{Proceedings of the European Conference on Computer Vision (ECCV)}}. \bibinfo{pages}{186--201}.
\newblock


\bibitem[Yue et~al\mbox{.}(2016)]%
        {yue2016deep}
\bibfield{author}{\bibinfo{person}{Cao Yue}, \bibinfo{person}{M Long}, \bibinfo{person}{J Wang}, \bibinfo{person}{Zhu Han}, {and} \bibinfo{person}{Q Wen}.} \bibinfo{year}{2016}\natexlab{}.
\newblock \showarticletitle{Deep quantization network for efficient image retrieval}. In \bibinfo{booktitle}{\emph{Proc. 13th AAAI Conf. Artif. Intell.}} \bibinfo{pages}{3457--3463}.
\newblock


\bibitem[Zhang et~al\mbox{.}(2017)]%
        {zhifei2017cvpr}
\bibfield{author}{\bibinfo{person}{Zhifei Zhang}, \bibinfo{person}{Yang Song}, {and} \bibinfo{person}{Hairong Qi}.} \bibinfo{year}{2017}\natexlab{}.
\newblock \showarticletitle{Age progression/regression by conditional adversarial autoencoder}. In \bibinfo{booktitle}{\emph{Proceedings of the IEEE conference on computer vision and pattern recognition}}. \bibinfo{pages}{5810--5818}.
\newblock


\bibitem[Zhou et~al\mbox{.}(2016a)]%
        {ZhouNZWWZ16}
\bibfield{author}{\bibinfo{person}{Shuchang Zhou}, \bibinfo{person}{Zekun Ni}, \bibinfo{person}{Xinyu Zhou}, \bibinfo{person}{He Wen}, \bibinfo{person}{Yuxin Wu}, {and} \bibinfo{person}{Yuheng Zou}.} \bibinfo{year}{2016}\natexlab{a}.
\newblock \showarticletitle{DoReFa-Net: Training Low Bitwidth Convolutional Neural Networks with Low Bitwidth Gradients}.
\newblock \bibinfo{journal}{\emph{CoRR}}  \bibinfo{volume}{abs/1606.06160} (\bibinfo{year}{2016}).
\newblock
\showeprint[arXiv]{1606.06160}
\urldef\tempurl%
\url{http://arxiv.org/abs/1606.06160}
\showURL{%
\tempurl}


\bibitem[Zhou et~al\mbox{.}(2016b)]%
        {zhou2016dorefa}
\bibfield{author}{\bibinfo{person}{Shuchang Zhou}, \bibinfo{person}{Yuxin Wu}, \bibinfo{person}{Zekun Ni}, \bibinfo{person}{Xinyu Zhou}, \bibinfo{person}{He Wen}, {and} \bibinfo{person}{Yuheng Zou}.} \bibinfo{year}{2016}\natexlab{b}.
\newblock \showarticletitle{Dorefa-net: Training low bitwidth convolutional neural networks with low bitwidth gradients}.
\newblock \bibinfo{journal}{\emph{arXiv preprint arXiv:1606.06160}} (\bibinfo{year}{2016}).
\newblock


\bibitem[Zhu et~al\mbox{.}(2018)]%
        {zhu2018euphrates}
\bibfield{author}{\bibinfo{person}{Yuhao Zhu}, \bibinfo{person}{Anand Samajdar}, \bibinfo{person}{Matthew Mattina}, {and} \bibinfo{person}{Paul Whatmough}.} \bibinfo{year}{2018}\natexlab{}.
\newblock \showarticletitle{Euphrates: Algorithm-SoC Co-Design for Low-Power Mobile Continuous Vision}. In \bibinfo{booktitle}{\emph{2018 ACM/IEEE 45th Annual International Symposium on Computer Architecture (ISCA)}}. IEEE, \bibinfo{pages}{547--560}.
\newblock


\end{thebibliography}

\end{document}